\pdfoutput=1

\documentclass[11pt]{article}
\usepackage{enumitem}
\usepackage[final]{coling}
\usepackage{bbding}
\usepackage{graphicx}
\usepackage{booktabs}
\usepackage{array}
\usepackage{longtable}
\usepackage{lipsum}

\usepackage{array}
\usepackage{multirow}
\usepackage{graphicx}
\usepackage{longtable}
\usepackage{algorithm,algorithmic}
\usepackage{hyperref}
\usepackage{times}
\usepackage{svg}
\usepackage{latexsym}
\usepackage{multirow}
\usepackage{threeparttable}
\usepackage{amsmath}
\usepackage{graphicx}
\usepackage{booktabs}
\usepackage{adjustbox}
\usepackage[T1]{fontenc}
\usepackage{makecell}
\usepackage[utf8]{inputenc}
\usepackage{svg}
\usepackage{microtype}
\setlist{nosep}
\usepackage{inconsolata}

\usepackage{multirow}
\usepackage{graphicx}
\RequirePackage{booktabs}
\title{GraCoRe: Benchmarking Graph Comprehension and Complex Reasoning in Large Language Models}

\author{
 \textbf{Zike Yuan\textsuperscript{1,2}},
 \textbf{Ming Liu\textsuperscript{1,2}},
 \textbf{Hui Wang\textsuperscript{2,*}},
 \textbf{Bing Qin\textsuperscript{1,2,*}}
\\
 \textsuperscript{1}Harbin Institute of Technology, Shenzhen, China,\\
 \textsuperscript{2}Peng Cheng Laboratory, Shenzhen, China\\
 \texttt{\{yuanzk,wangh06\}@pcl.ac.cn}\\
 \texttt{\{mliu,qinb\}@ir.hit.edu.cn}
\\
}

\begin{document}
\maketitle

\begin{abstract}
Evaluating the graph comprehension and reasoning abilities of Large Language Models (LLMs) is challenging and often incomplete. Existing benchmarks focus primarily on pure graph understanding, lacking a comprehensive evaluation across all graph types and detailed capability definitions. This paper presents GraCoRe, a benchmark for systematically assessing LLMs' graph comprehension and reasoning. GraCoRe uses a three-tier hierarchical taxonomy to categorize and test models on pure graph and heterogeneous graphs, subdividing capabilities into 10 distinct areas tested through 19 tasks. Our benchmark includes 11 datasets with 5,140 graphs of varying complexity. We evaluate four closed-source and eight open-source LLMs, conducting thorough analyses from both ability and task perspectives. Key findings reveal that OpenAI o1 model has amazing comprehension and reasoning capabilities, semantic enrichment enhances reasoning performance, node ordering impacts task success, and the ability to process longer texts does not necessarily improve graph comprehension or reasoning.GraCoRe is open-sourced at \href{https://github.com/ZIKEYUAN/GraCoRe}{https://github.com/ZIKEYUAN/GraCoRe}.
\end{abstract}
\renewcommand{\thefootnote}{}
\footnotetext{*  B. Qin and H. Wang are corresponding authors.}
\section{Introduction}

Graph understanding and complex reasoning are crucial capabilities of Large Language Models (LLMs), supporting applications in areas like social network analysis, drug discovery, recommendation systems, and spatiotemporal prediction \citep{Inro2}. These abilities are particularly important for advancing Artificial General Intelligence (AGI) \citep{Inro3}. Graph-structured data mainly consists of homogeneous and heterogeneous graphs. Research on homogeneous graphs often addresses specific structural issues, such as protein-protein interaction prediction \citep{protein}. Unlike homogeneous graphs, pure graphs are simpler, lacking node and edge attributes, as seen in graph-theoretic problems \citep{borgatti2006graph}. For heterogeneous graphs, tasks leverage rich semantic information, like knowledge graph reasoning. However, enabling LLMs to effectively parse, understand, and reason with graph-structured data remains a significant challenge.
\begin{figure}[t]
  \includegraphics[width=\columnwidth]{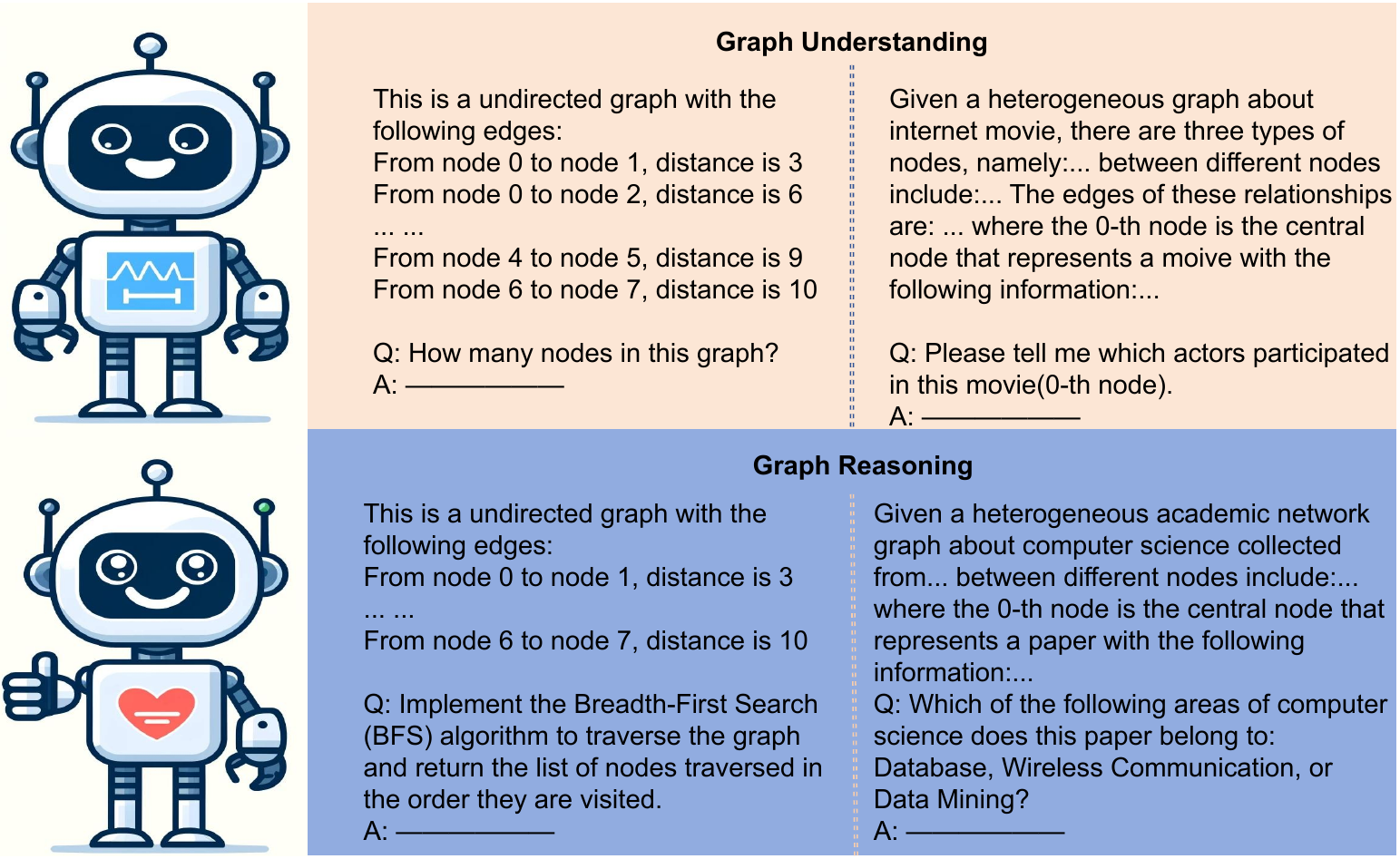}
  \caption{GraCoRe encompasses two overarching abilities and 19 distinct tasks within LLM on graph scenarios, facilitating a granular benchmarking from basic perceptivity to advanced interactivity.}
     \vspace{-20pt}
  \label{fig:1}
\end{figure}

Firstly, the reasoning and comprehension abilities of current LLMs degrade significantly when processing complex graph-structured data with numerous nodes and edges. This decline is partly due to challenges in handling lengthy textual inputs that describe such data. Long texts increase computational burden and introduce noise, reducing the models' ability to capture essential details. Secondly, textual descriptions often involve complex entity relationships and abstract concepts, requiring models to understand explicit information and infer implicit connections. Current research primarily focuses on direct mappings from graph structure to answers \citep{NLgraph-6}, overlooking deeper reasoning capabilities. The absence of clear definitions and evaluation standards for graph reasoning highlights the need for a comprehensive benchmark to evaluate these abilities.

Previous research has introduced several benchmarks to evaluate LLMs' understanding and reasoning on graphs, but these benchmarks have notable limitations. Firstly, there is a \textbf{limited generalization} issue: most benchmarks test either pure or heterogeneous graphs separately, lacking a unified evaluation across both types. Additionally, there is a \textbf{lack of clear definition of model capabilities}: existing benchmarks are task-driven and fail to assess LLMs' specific abilities with graph data. Therefore, more comprehensive, ability-based benchmarks are needed to evaluate understanding and reasoning on graph-structured data. Lastly, there is \textbf{insufficient diversity in model types and tasks}: current benchmarks do not clearly classify tasks or test a wide range of models.

To address these challenges, we propose the GraCoRe benchmark, as shown in Figure \ref{fig:1}, which aims to explore and evaluate the understanding and reasoning capabilities of mainstream LLMs. We have designed a \textbf{three-tier hierarchical ability taxonomy} that includes both capability-based and dataset-based categories. This taxonomy meticulously defines the model's capabilities and ensures greater generalization in the testing range. Regarding model types and task divisions, our benchmark tests multiple existing closed-source and open-source models, and divides tasks into multiple dimensions based on model capabilities. In Appendix \ref{sec:A}, we compare in detail the differences between the existing benchmark and GraCoRe, as well as our advantages.

\begin{figure}[t]
  \includegraphics[width=\columnwidth]{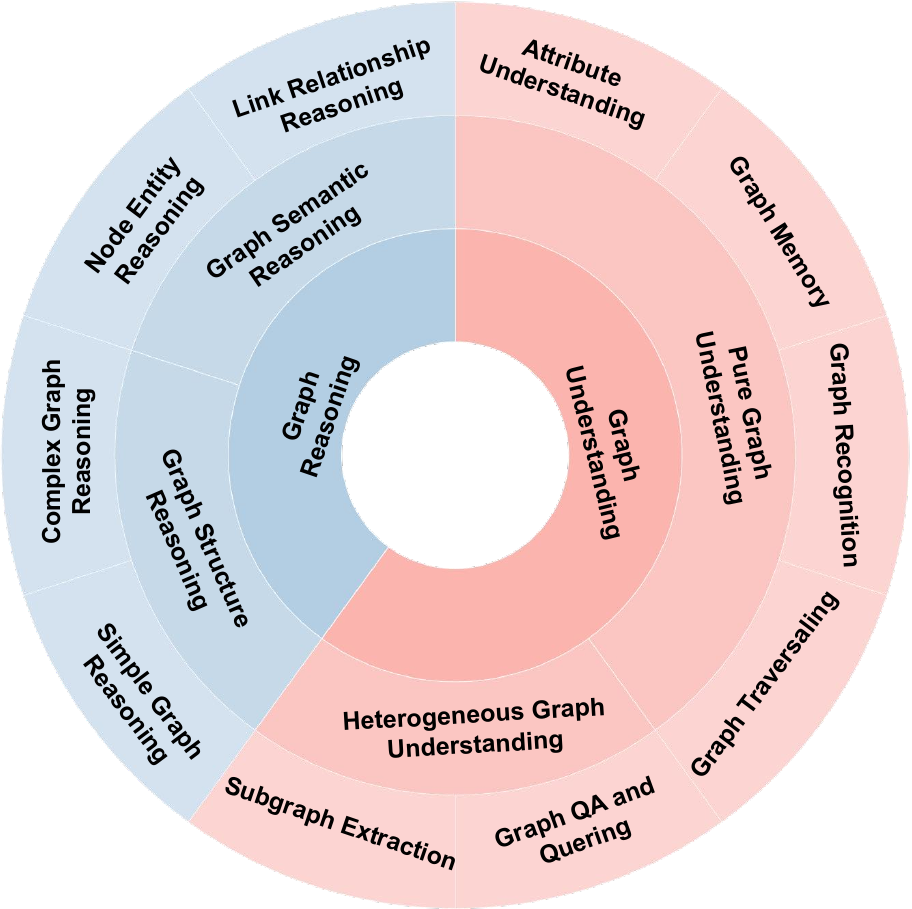}
  \caption{Our three-tier hierarchical ability taxonomy.}
  \label{fig:2}
     \vspace{-22pt}
\end{figure}

Figure \ref{fig:2} presents the framework of our ability taxonomy. The first layer highlights two core capabilities: \textbf{graph understanding} and \textbf{graph reasoning}. Graph understanding reflects the model's ability to comprehend nodes, edges, and graph structure within the context, while graph reasoning builds on this, focusing on inferring implicit information from graph-structured data. The second layer categorizes LLM capabilities into four types based on data types. In the third layer, these are further divided into 10 distinct capabilities, evaluated through 19 tasks. This taxonomy provides multi-level evaluation, enabling detailed identification of model weaknesses. Each task is designed with specific prompts, structured by predefined rules to textualize graph information. GraCoRe includes 11 datasets with 5,140 graphs, with graph complexity controlled by factors such as size and network sparsity.

 We conducted extensive experiments on GraCoRe to evaluate the graph understanding and reasoning abilities of LLMs, including four closed-source and eight open-source models. Key findings include:
\begin{itemize}[left=0pt]
	\item Graph reasoning is a major weakness in current LLMs, with most models unable to balance understanding and reasoning. OpenAI o1 model has amazing understanding and reasoning capabilities.
	\item LLMs perform better on graph reasoning tasks with semantic information than on purely structural tasks, showing that textual context enhances reasoning.
	\item Model performance is sensitive to node ordering in textual graph data, where ordered naming improves results.
	\item The ability to handle longer text does not impact performance, regardless of graph complexity or description length.
\end{itemize}

\section{Related Work}
\textbf{LLMs for Graph}
Recent advancements in LLMs for graph tasks includes several notable contributions. \citep{suvery} categorizes these tasks into three types: Enhancer, Predictor, and Alignment. \citep{2} provides a forward-looking roadmap for the unification of LLMs and Knowledge Graphs (KGs).\citep{3} proposes an end-to-end method for solving graph-related problems. Furthermore, \citep{4} investigates methods to improve the zero-shot reasoning ability of LLMs over structured data in a unified manner. \citep{5} explores the graph generation capabilities of LLMs through systematic task designs and extensive experiments.

\noindent\textbf{Benchmarks for LLMs on Graph} 
Most benchmarks for evaluating LLMs on graph tasks are based on task testing. NLGraph\citep{NLgraph-6} introduced a simple test dataset for eight graph tasks, while GPT4Graph\citep{gpt4graph-7} tested LLM capabilities on semantic tasks. \citep{8} assessed the capabilities of four LLMs in graph data analysis. \citep{talkgraph-9} proposed a method for describing graph data in text form. \citep{10} designed a hint method specifically for graph tasks. GraphInstruct\citep{11} provided diverse processes and steps for graph data generation. HiGPT\citep{12} proposed an evaluation method for heterogeneous graphs, and VisionGraph\citep{13} assessed the capabilities of LLMs on image graphs.





\renewcommand{\arraystretch}{0.8} 
\setlength{\tabcolsep}{4pt} 
\begin{table*}[ht]
\small
	\setlength{\tabcolsep}{0.pt}
    \centering
    
    \begin{tabular}{@{}>{\raggedright\arraybackslash}p{3.5cm} >{\centering\arraybackslash}p{1.5cm} >{\raggedright\arraybackslash}p{11cm}@{}}
        \toprule
        \textbf{Task} & \textbf{Abbr.} & \textbf{Description} \\ 
        \midrule
        Node Number & NN & Calculate the total number of nodes in a graph. \\ 
        \midrule
        Average Degree & AD & Calculate the average degree of the nodes in a graph. \\ 
        \midrule
        Connectivity Test & CT & Determine if the graph is connected, meaning there is a path between any two nodes. \\ 
        \midrule
        Matrix Similarity & MS & Evaluate the similarity between two adjacency matrices created from LLM and target graph. \\ 
        \midrule
        Tree Recognition & TR & Identify if a graph is binary tree. \\ 
        \midrule
        Bipartite Recognition & BR & Identify if a graph is bipartite, meaning its nodes can be divided into two disjoint sets such that no two nodes within the same set are adjacent. \\ 
        \midrule
        Breadth First Search & BFS & Perform a breadth-first traversal starting from a given node. \\ 
        \midrule
        Neighborhood Query & NQ & Query all nodes that are neighbors of a specified node. \\ 
        \midrule
        Relationship Query & RQ & Query for specific relationships between nodes in the graph. \\ 
        \midrule
        Relation Number & RN & Count the number of relationship types in the graph. \\ 
        \midrule
        Subgraph Extraction & SE & Extract a subgraph based on specified criteria or nodes. \\ 
        \midrule
        Shortest Path & SP & Find the shortest path between two nodes in a graph. \\ 
        \midrule
        Maximum Flow & MF & Calculate the maximum flow in a flow network. \\ 
        \midrule
        Eulerian Path & EP & Determine if there is a path that visits every edge exactly once. \\ 
        \midrule
        Hamiltonian Cycle & HC & Determine if there is a cycle that visits every node exactly once and back start node. \\ 
        \midrule
        Traveling Salesman Problem & TSP & Find the shortest possible route that visits each node exactly once and returns to the origin node. \\ 
        \midrule
        Graph Coloring & GC & Assign colors to the nodes of the graph so that no two adjacent nodes share the same color. \\ 
        \midrule
        Node Classification & NC & Classify nodes into predefined categories based on their attributes or graph structure. \\ 
        \midrule
        Link Prediction & LP & Predict whether a link (edge) exists between two nodes in the graph. \\ 
        \bottomrule
    \end{tabular}
    \caption{ The 19 tasks for LLMs on graph within GraCoRe.}
    \label{tab:1}
         \vspace{-15pt}
\end{table*}
\section{GraCoRe}

This section begins by describing a three-tier hierarchical taxonomy for LLMs on graph data. Next, we explain the methodology used to collect the dataset. Finally, we present an analysis of the dataset's statistics.

\subsection{Hierarchical Ability Taxonomy}
After analyzing \citep{mtbench} evaluation of the LLMs in the multi-round dialogue task, we have developed a hierarchical taxonomy for classifying LLMs capabilities on graphs, essential for their evaluation. The taxonomy is structured into three levels, encompassing 19 tasks within 10 sub-capabilities. Table \ref{tab:1} summarizes each task with a brief description. This section elaborates on the three levels and their corresponding tasks, with detailed examples provided in the Appendix \ref{sec:B}.
\begin{table*}[t]
		\setlength{\tabcolsep}{1pt}
    \footnotesize
    \centering
    \begin{tabular}{@{}p{2.5cm} p{3cm} p{1.5cm} p{1.8cm} p{1.5cm} p{1cm} p{1cm} p{1cm}@{}}
        \toprule
        \textbf{Benchmark} & \textbf{Graph Type} & \textbf{\#Graph} & \textbf{Average Node} & \textbf{Average Edge} & \textbf{\#Node Type} & \textbf{\#Edge Type} & \textbf{\#Task} \\ 
        \midrule
        \multirow{10}{*}{PureGra} & Bipartite Graph & 460 & 19 & 49 & 1 & 1 & 5 \\ 
        & Tree Structure Graph & 460 & 19 & 18 & 1 & 1 & 5 \\ 
        & Graph Traversal Graph & 460 & 19 & 53 & 1 & 1 & 5 \\ 
        & Shortest Path Graph & 460 & 19 & 53 & 1 & 1 & 5 \\ 
        & Max Flow Graph & 460 & 19 & 106 & 1 & 1 & 1 \\ 
        & Graph Coloring & 460 & 19 & 53 & 1 & 1 & 1 \\ 
        & Hamiltonian Graph & 460 & 19 & 72 & 1 & 1 & 1 \\ 
        & TSP Graph & 460 & 19 & 193 & 1 & 1 & 1 \\ 
        & Eulerian Graph & 460 & 19 & 104 & 1 & 1 & 1 \\ 
        \midrule
        \multirow{2}{*}{\makecell{HeterGra}} & IMDBText & 500 & 27 & 30 & 3 & 4 & 4 \\ 
        & ACMText & 500 & 158 & 171 & 4 & 8 & 2 \\ 
        \midrule
        
        \multirow{1}{*}{\textbf{GraCoRe}} & /\ & \textbf{5140} & / & / & / & / & \textbf{19} \\ 
        
        \bottomrule
    \end{tabular}
    \caption{Benchmark Graph Statistics.}
    \label{tab:2}
    \vspace{-15pt}
\end{table*}
\subsubsection{Graph understanding}
Understanding graph structure necessitates LLMs capable of accurately answering questions about the graph's basic properties and reconstructing its structural information from extensive text descriptions. This involves two core capabilities:

\noindent \textbf{Pure graph understanding:} Pure graphs refer to graph data containing only structural information, representing the simplest form of graph data\citep{jin2023large}. The structural information of such graphs can be encapsulated in an adjacency matrix. Consequently, research on pure graphs often emphasizes the structural information. Assessing the ability of LLMs to understand pure graphs should thus focus on their capacity to comprehend structural information. To this end, I have identified four sub-capabilities:
\begin{itemize}[left=0pt]
    \item \textbf{Pure Graph Attribute Understanding:} The most intuitive measure of understanding graph structure information is the correct comprehension of its basic attributes, such as the number of nodes, average degree, and performance on several sub-tasks related to node connectivity.
    \item \textbf{Pure Graph Memory:} This requires the large language model to reconstruct the input graph structure data, testing its memory capacity. This is specifically evaluated by the similarity score\citep{lahitani2016cosine} of the reconstructed matrix.
    \item \textbf{Pure Graph Recognition:} For graph data with different structures, the model must be able to recognize and distinguish them. This study uses bipartite graphs and tree graphs to evaluate this capability.
    \item \textbf{Graph Traversal:} Traversal is fundamental to solving many graph theory reasoning problems. The model's performance in reasoning tasks is influenced by its traversal capability. This study primarily tests whether the model can traverse a graph using Breadth-First Search (BFS)\citep{BFSbeamer2013direction}.
\end{itemize}

\noindent\textbf{Heterogeneous graph understanding:} Unlike pure graphs, heterogeneous graphs often contain rich semantic information, with much of the data collected from real-world scenarios. Consequently, understanding heterogeneous graphs typically involves grasping their semantic information, whereas understanding their structural information is less critical. We have refined this into two sub-capabilities:
\begin{itemize}[left=0pt]
    \item \textbf{Graph QA and Querying:} Given the rich semantic information in heterogeneous graphs, the ability of large language models to perform question-answering and querying is crucial. This capability includes three sub-tasks: querying neighbor nodes, answering questions about node relationships, and querying the number of relationships.
    \item \textbf{Subgraph Extraction:} This pertains to the overall understanding of relationships and nodes within heterogeneous graphs, assessing the model's ability to extract relevant subgraphs.
\end{itemize}

\subsubsection{Graph reasoning}
Based on the graph understanding capabilities of LLMs, their graph reasoning abilities are also worth exploring. This ability requires the model to infer hidden information from known graph data. It includes the following two core capabilities:

\noindent\textbf{Graph structure reasoning:} Graph structural information reasoning requires large language models to understand the nodes, edges, and their connections to infer the overall structural characteristics of the graph or the structural patterns of specific subgraphs. For example, the model should be able to identify cycles, paths, tree structures, and hierarchical structures within the graph, and use these structural features for further reasoning. Tasks related to structural reasoning are primarily focused on graph theory problems. We classify the complexity of these problems into two categories:\textbf{ Simple Graph Theory Problems Reasoning} and \textbf{Complex Graph Theory Problems Reasoning}. The classification criterion is the time complexity of the corresponding algorithms. Simple graph theory problems are solvable in polynomial time, while complex graph theory problems are NP-complete, requiring significantly more time to solve. This classification tests the model's capability in reasoning about graph theory problems. We have selected three representative problems for testing within each of these categories.

\noindent\textbf{Graph semantic reasoning:}  Unlike structure-based reasoning, semantic information reasoning in graphs requires large language models to deeply understand the semantic meanings of nodes and edges, and to reason based on this semantic information. This involves modeling and reasoning about entities, relationships, and their interactions within the graph. Based on these tasks, we subdivide the semantic reasoning capabilities of large language models into \textbf{Node Entity Reasoning} and \textbf{Link Relationship Reasoning}. Corresponding tasks include node classification and link prediction. Previous studies have primarily addressed these two problems using graph neural networks (GNNs) such as GCN\citep{GCN} and GraphSAGE\citep{sage}. These methods typically require large structured graph data for training and cannot directly utilize graph data containing text information for inference. Consequently, investigating the use of large models for semantic reasoning is highly significant. This research will explore whether text enhancement impacts the performance of LLMs on these two tasks

\subsection{Data Collection}
We first divide the dataset into pure graphs and heterogeneous graphs to test the capabilities of large language models on these two types of datasets. For pure graphs, we customize unique data generation prompts based on the specific characteristics of each task, generating corresponding graph structure data using manually set rules. The scale of the graph is defined by the number of nodes and the sparsity of the network, ensuring that the generated data meets the specific needs of each task. For heterogeneous graph data, we use the ACM\citep{ACM} and IMDB\citep{IDBM} datasets, converting them into text-based graph data. These are constructed according to a manually specified graph structure description framework, and unique prompts are designed for each task to build the dataset.

After generating the benchmark datasets, we also designed specific few-shot prompts for each task to test the model's capabilities. These prompt datasets will be included in the benchmark data to provide additional testing options. Finally, we will provide a standard answer for each task and filter out graph data that does not meet the corresponding task requirements.
\begin{table*}[]
	\centering
	\scriptsize
	\setlength{\extrarowheight}{2.5pt} 
 \resizebox{\linewidth}{!}{

	\begin{tabular}{c|ccccccc|cccc}
		\hline
		\multirow{3}{*}{\textbf{Model}} & \multicolumn{7}{c|}{\textbf{Pure Graph}}                                                                                                                                                                                                                                                                                                                                        & \multicolumn{4}{c}{\textbf{Heterogeneous Graph}}                                                                                                                                             \\ \cline{2-12} 
		& \multicolumn{3}{c|}{\textbf{\begin{tabular}[c]{@{}c@{}}Attribute\\  Understanding\end{tabular}}}                   & \multicolumn{1}{c|}{\textbf{\begin{tabular}[c]{@{}c@{}}Graph \\ Memory\end{tabular}}} & \multicolumn{2}{c|}{\textbf{\begin{tabular}[c]{@{}c@{}}Graph \\ Recognition\end{tabular}}} & \textbf{\begin{tabular}[c]{@{}c@{}}Grap \\ Traversaling\end{tabular}} & \multicolumn{3}{c|}{\textbf{\begin{tabular}[c]{@{}c@{}}Graph QA \\ and Quering\end{tabular}}}                      & \textbf{\begin{tabular}[c]{@{}c@{}}Subgraph \\ Extraction\end{tabular}} \\ \cline{2-12} 
		& \multicolumn{1}{c|}{\textbf{NN}}     & \multicolumn{1}{c|}{\textbf{AD}}     & \multicolumn{1}{c|}{\textbf{CT}}     & \multicolumn{1}{c|}{\textbf{MS}}                                                      & \multicolumn{1}{c|}{\textbf{TR}}              & \multicolumn{1}{c|}{\textbf{BR}}           & \textbf{BFS}                                                          & \multicolumn{1}{c|}{\textbf{NQ}}     & \multicolumn{1}{c|}{\textbf{RQ}}     & \multicolumn{1}{c|}{\textbf{RN}}     & \textbf{SE}                                                             \\ \hline
		\textit{OpenAI o1}         & \multicolumn{1}{c|}{57.836}          & \multicolumn{1}{c|}{\textbf{81.591}}  & \multicolumn{1}{c|}{\textbf{58.491}}                  & \multicolumn{1}{c|}{\textbf{58.959}}   & \multicolumn{1}{c|}{\textbf{96.545}}             & \multicolumn{1}{c|}{\textbf{71.257}}                  & \multicolumn{1}{c|}{\textbf{66.263}}    & \multicolumn{1}{c|}{\textbf{72.029}}  & \multicolumn{1}{c|}{\textbf{58.731}}    & \multicolumn{1}{c|}{62.463}          & \textbf{75.531}     \\
\textit{GPT-4o}          & \multicolumn{1}{c|}{\textbf{58.978}} & \multicolumn{1}{c|}{59.195}           & \multicolumn{1}{c|}{41.355}                           & \multicolumn{1}{c|}{58.906}            & \multicolumn{1}{c|}{50.275}                      & \multicolumn{1}{c|}{41.450}                           & \multicolumn{1}{c|}{55.864}             & \multicolumn{1}{c|}{57.564}           & \multicolumn{1}{c|}{53.095}             & \multicolumn{1}{c|}{60.99}           & 60.042              \\
\textit{GPT-4}           & \multicolumn{1}{c|}{58.887}          & \multicolumn{1}{c|}{52.842}           & \multicolumn{1}{c|}{56.054}                           & \multicolumn{1}{c|}{58.853}            & \multicolumn{1}{c|}{43.737}                      & \multicolumn{1}{c|}{26.832}                           & \multicolumn{1}{c|}{55.349}             & \multicolumn{1}{c|}{63.648}           & \multicolumn{1}{c|}{52.898}             & \multicolumn{1}{c|}{\textbf{63.128}} & 60.763              \\
\textit{GPT-3.5}         & \multicolumn{1}{c|}{\textbf{58.978}} & \multicolumn{1}{c|}{42.041}           & \multicolumn{1}{c|}{37.166}                           & \multicolumn{1}{c|}{56.521}            & \multicolumn{1}{c|}{28.775}                      & \multicolumn{1}{c|}{60.293}                           & \multicolumn{1}{c|}{60.949}             & \multicolumn{1}{c|}{41.671}           & \multicolumn{1}{c|}{26.302}             & \multicolumn{1}{c|}{45.879}          & 43.594              \\
\textit{Llama3.1-ins-8b} & \multicolumn{1}{c|}{45.869}          & \multicolumn{1}{c|}{57.845}           & \multicolumn{1}{c|}{35.643}                           & \multicolumn{1}{c|}{36.175}            & \multicolumn{1}{c|}{33.679}                      & \multicolumn{1}{c|}{27.517}                           & \multicolumn{1}{c|}{43.522}             & \multicolumn{1}{c|}{52.908}           & \multicolumn{1}{c|}{50.327}             & \multicolumn{1}{c|}{34.094}          & 44.614              \\
\textit{Llama3-ins-8b}   & \multicolumn{1}{c|}{54.228}          & \multicolumn{1}{c|}{26.077}           & \multicolumn{1}{c|}{26.275}                           & \multicolumn{1}{c|}{44.918}            & \multicolumn{1}{c|}{41.851}                      & \multicolumn{1}{c|}{25.576}                           & \multicolumn{1}{c|}{60.835}             & \multicolumn{1}{c|}{29.628}           & \multicolumn{1}{c|}{52.502}             & \multicolumn{1}{c|}{57.996}          & 33.748              \\
\textit{Qwen2-7b-ins}    & \multicolumn{1}{c|}{53.634}          & \multicolumn{1}{c|}{16.229}           & \multicolumn{1}{c|}{56.968}                           & \multicolumn{1}{c|}{52.283}            & \multicolumn{1}{c|}{36.445}                      & \multicolumn{1}{c|}{47.959}                           & \multicolumn{1}{c|}{29.237}             & \multicolumn{1}{c|}{39.498}           & \multicolumn{1}{c|}{52.502}             & \multicolumn{1}{c|}{19.364}          & 55.360              \\
\textit{Llama2-7b-chat}  & \multicolumn{1}{c|}{17.048}          & \multicolumn{1}{c|}{35.766}           & \multicolumn{1}{c|}{53.084}                           & \multicolumn{1}{c|}{38.613}            & \multicolumn{1}{c|}{33.930}                      & \multicolumn{1}{c|}{25.576}                           & \multicolumn{1}{c|}{38.836}             & \multicolumn{1}{c|}{29.007}           & \multicolumn{1}{c|}{0.991}              & \multicolumn{1}{c|}{35.473}          & 28.946              \\
\textit{Vicuna-v1.5-16k} & \multicolumn{1}{c|}{31.755}          & \multicolumn{1}{c|}{29.175}           & \multicolumn{1}{c|}{0.000}                                & \multicolumn{1}{c|}{37.818}            & \multicolumn{1}{c|}{30.410}                      & \multicolumn{1}{c|}{40.993}                           & \multicolumn{1}{c|}{23.808}             & \multicolumn{1}{c|}{34.718}           & \multicolumn{1}{c|}{51.118}             & \multicolumn{1}{c|}{53.720}          & 29.486              \\
\textit{Chatglm3-6b}     & \multicolumn{1}{c|}{23.442}          & \multicolumn{1}{c|}{33.066}           & \multicolumn{1}{c|}{55.140}                           & \multicolumn{1}{c|}{37.659}            & \multicolumn{1}{c|}{33.050}                      & \multicolumn{1}{c|}{25.576}                           & \multicolumn{1}{c|}{28.151}             & \multicolumn{1}{c|}{13.735}           & \multicolumn{1}{c|}{21.061}             & \multicolumn{1}{c|}{15.610}          & 17.300              \\
\textit{Chatglm2-32k-7b} & \multicolumn{1}{c|}{13.987}          & \multicolumn{1}{c|}{23.377}           & \multicolumn{1}{c|}{17.974}                           & \multicolumn{1}{c|}{6.239}             & \multicolumn{1}{c|}{35.816}                      & \multicolumn{1}{c|}{74.683}                           & \multicolumn{1}{c|}{12.895}             & \multicolumn{1}{c|}{46.576}           & \multicolumn{1}{c|}{21.061}             & \multicolumn{1}{c|}{15.610}          & 24.744              \\

\textit{Vicuna-v1.5-7b}  & \multicolumn{1}{c|}{18.646}          & \multicolumn{1}{c|}{36.084}           & \multicolumn{1}{c|}{55.140}                           & \multicolumn{1}{c|}{6.345}             & \multicolumn{1}{c|}{28.775}                      & \multicolumn{1}{c|}{25.576}                           & \multicolumn{1}{c|}{17.580}             & \multicolumn{1}{c|}{12.307}           & \multicolumn{1}{c|}{52.700}             & \multicolumn{1}{c|}{28.962}          & 19.161                                                                             \\ \hline
	\end{tabular}}
     \caption{Standardized performance of graph understanding.}
    \label{tab:3}
\end{table*}

\begin{table*}[]
	\centering
	\scriptsize
	\setlength{\extrarowheight}{2.5pt} 
	\resizebox{\linewidth}{!}{

	\begin{tabular}{c|cccccc|cc|c|c}
		\hline
		\multirow{3}{*}{\textbf{Model}} & \multicolumn{6}{c|}{\textbf{Graph Structure Reasoning}}                                                                                                                                                                     & \multicolumn{2}{c|}{\textbf{Graph Semantic Reasoning}}                                                                                                                           & \multirow{3}{*}{\textbf{Average $z$}} & \multirow{3}{*}{\textbf{Total Score}} \\ \cline{2-9}
		& \multicolumn{3}{c|}{\textbf{\begin{tabular}[c]{@{}c@{}}Simple Graph \\ Theory Problems\end{tabular}}}              & \multicolumn{3}{c|}{\textbf{\begin{tabular}[c]{@{}c@{}}Complex Graph \\ Theory Problems\end{tabular}}} & \multicolumn{1}{c|}{\textbf{\begin{tabular}[c]{@{}c@{}}Node Entity \\ Reasoning\end{tabular}}} & \textbf{\begin{tabular}[c]{@{}c@{}}Link Relationship \\ Reasoning\end{tabular}} &                                     &                                       \\ \cline{2-9}
		& \multicolumn{1}{c|}{\textbf{SP}}     & \multicolumn{1}{c|}{\textbf{MF}}     & \multicolumn{1}{c|}{\textbf{EP}}     & \multicolumn{1}{c|}{\textbf{HC}}        & \multicolumn{1}{c|}{\textbf{TSP}}       & \textbf{GC}        & \multicolumn{1}{c|}{\textbf{NC}}                                                               & \textbf{LP}                                                                     &                                     &                                       \\ \hline
		\textit{OpenAI o1}         & \multicolumn{1}{c|}{\textbf{95.320}} & \multicolumn{1}{c|}{\textbf{99.390}} & \multicolumn{1}{c|}{\textbf{71.313}} & \multicolumn{1}{c|}{\textbf{85.057}}  & \multicolumn{1}{c|}{\textbf{100.000}}   & \multicolumn{1}{c|}{\textbf{73.379}} & \multicolumn{1}{c|}{66.409}                & 53.236                      & \textbf{1.762} & \textbf{1403.799} \\
\textit{GPT-4o}          & \multicolumn{1}{c|}{52.665}          & \multicolumn{1}{c|}{35.019}          & \multicolumn{1}{c|}{61.671}          & \multicolumn{1}{c|}{41.405}           & \multicolumn{1}{c|}{37.647}             & \multicolumn{1}{c|}{37.860}          & \multicolumn{1}{c|}{\textbf{66.538}}       & \textbf{62.877}             & 0.601          & 993.397           \\
\textit{GPT-4}           & \multicolumn{1}{c|}{42.456}          & \multicolumn{1}{c|}{33.045}          & \multicolumn{1}{c|}{62.255}          & \multicolumn{1}{c|}{43.303}           & \multicolumn{1}{c|}{36.700}             & \multicolumn{1}{c|}{33.954}          & \multicolumn{1}{c|}{65.435}                & 57.641                      & 0.528          & 967.778           \\
\textit{GPT-3.5}         & \multicolumn{1}{c|}{32.852}          & \multicolumn{1}{c|}{33.834}          & \multicolumn{1}{c|}{60.867}          & \multicolumn{1}{c|}{51.788}           & \multicolumn{1}{c|}{35.872}             & \multicolumn{1}{c|}{45.581}          & \multicolumn{1}{c|}{57.908}                & 48.830                      & 0.251          & 869.702           \\
\textit{Llama3.1-ins-8b} & \multicolumn{1}{c|}{35.101}          & \multicolumn{1}{c|}{33.045}          & \multicolumn{1}{c|}{35.446}          & \multicolumn{1}{c|}{54.021}           & \multicolumn{1}{c|}{33.624}             & \multicolumn{1}{c|}{77.194}          & \multicolumn{1}{c|}{40.064}                & 50.908                      & 0.115          & 821.597           \\
\textit{Llama3-ins-8b}   & \multicolumn{1}{c|}{39.081}          & \multicolumn{1}{c|}{35.809}          & \multicolumn{1}{c|}{30.698}          & \multicolumn{1}{c|}{56.477}           & \multicolumn{1}{c|}{34.334}             & \multicolumn{1}{c|}{49.306}          & \multicolumn{1}{c|}{36.041}                & 55.563                      & 0.028          & 790.943           \\
\textit{Qwen2-7b-ins}    & \multicolumn{1}{c|}{37.005}          & \multicolumn{1}{c|}{37.389}          & \multicolumn{1}{c|}{39.537}          & \multicolumn{1}{c|}{37.498}           & \multicolumn{1}{c|}{38.238}             & \multicolumn{1}{c|}{28.957}          & \multicolumn{1}{c|}{36.625}                & 52.903                      & -0.038         & 767.630           \\
\textit{Llama2-7b-chat}  & \multicolumn{1}{c|}{36.486}          & \multicolumn{1}{c|}{40.153}          & \multicolumn{1}{c|}{19.594}          & \multicolumn{1}{c|}{24.882}           & \multicolumn{1}{c|}{34.570}             & \multicolumn{1}{c|}{41.039}          & \multicolumn{1}{c|}{27.994}                & 33.786                      & -0.524         & 595.774           \\
\textit{Vicuna-v1.5-16k} & \multicolumn{1}{c|}{34.236}          & \multicolumn{1}{c|}{32.255}          & \multicolumn{1}{c|}{20.179}          & \multicolumn{1}{c|}{24.659}           & \multicolumn{1}{c|}{34.334}             & \multicolumn{1}{c|}{16.512}          & \multicolumn{1}{c|}{12.746}                & 32.955                      & -0.595         & 570.875           \\
\textit{Chatglm3-6b}     & \multicolumn{1}{c|}{23.681}          & \multicolumn{1}{c|}{40.153}          & \multicolumn{1}{c|}{42.897}          & \multicolumn{1}{c|}{24.659}           & \multicolumn{1}{c|}{36.463}             & \multicolumn{1}{c|}{40.585}          & \multicolumn{1}{c|}{32.407}                & 21.734                      & -0.608         & 566.369           \\
\textit{Chatglm2-32k-7b} & \multicolumn{1}{c|}{26.103}          & \multicolumn{1}{c|}{40.153}          & \multicolumn{1}{c|}{28.141}          & \multicolumn{1}{c|}{24.659}           & \multicolumn{1}{c|}{35.872}             & \multicolumn{1}{c|}{22.871}          & \multicolumn{1}{c|}{20.013}                & 20.072                      & -0.765         & 510.845           \\

\textit{Vicuna-v1.5-7b}  & \multicolumn{1}{c|}{38.303}          & \multicolumn{1}{c|}{33.045}          & \multicolumn{1}{c|}{20.690}          & \multicolumn{1}{c|}{24.882}           & \multicolumn{1}{c|}{35.635}             & \multicolumn{1}{c|}{26.050}          & \multicolumn{1}{c|}{31.109}                & 2.783                       & -0.756         & 513.774                                  \\ \hline
	\end{tabular}}
      \caption{Standardized performance of graph reasoning.}
    \label{tab:4}
    \vspace{-15pt}
\end{table*}
\subsection{Data Statistics}

Table \ref{tab:2} shows several key statistics of our GraCoRe benchmark. We categorized the dataset based on graph structure into two main datasets: PureGra and HeterGra. Each main dataset contains multiple sub-datasets used for corresponding task testing. In total, there are 19 tasks with up to 5,140 graphs. Detailed statistics for each task can be found in the Appendix \ref{sec:C}.

For pure graph data, the datasets include graphs with 8 to 30 nodes, with 20 test graphs per dataset. This design is intended to assess the impact of graph complexity on the performance of large language models. For heterogeneous graph data, we divided them into IMDBText and ACMText datasets. The ACMText dataset is more complex and extensive, containing more semantic information than the IMDBText dataset. Therefore, the ACMText dataset is primarily used for complex reasoning tasks, including node classification and edge prediction.

GraCoRe is the first benchmark specifically focused on the fine-grained understanding and reasoning capabilities of large language models on graph data.



\begin{figure*}[t]
  \includegraphics[width=0.5\linewidth]{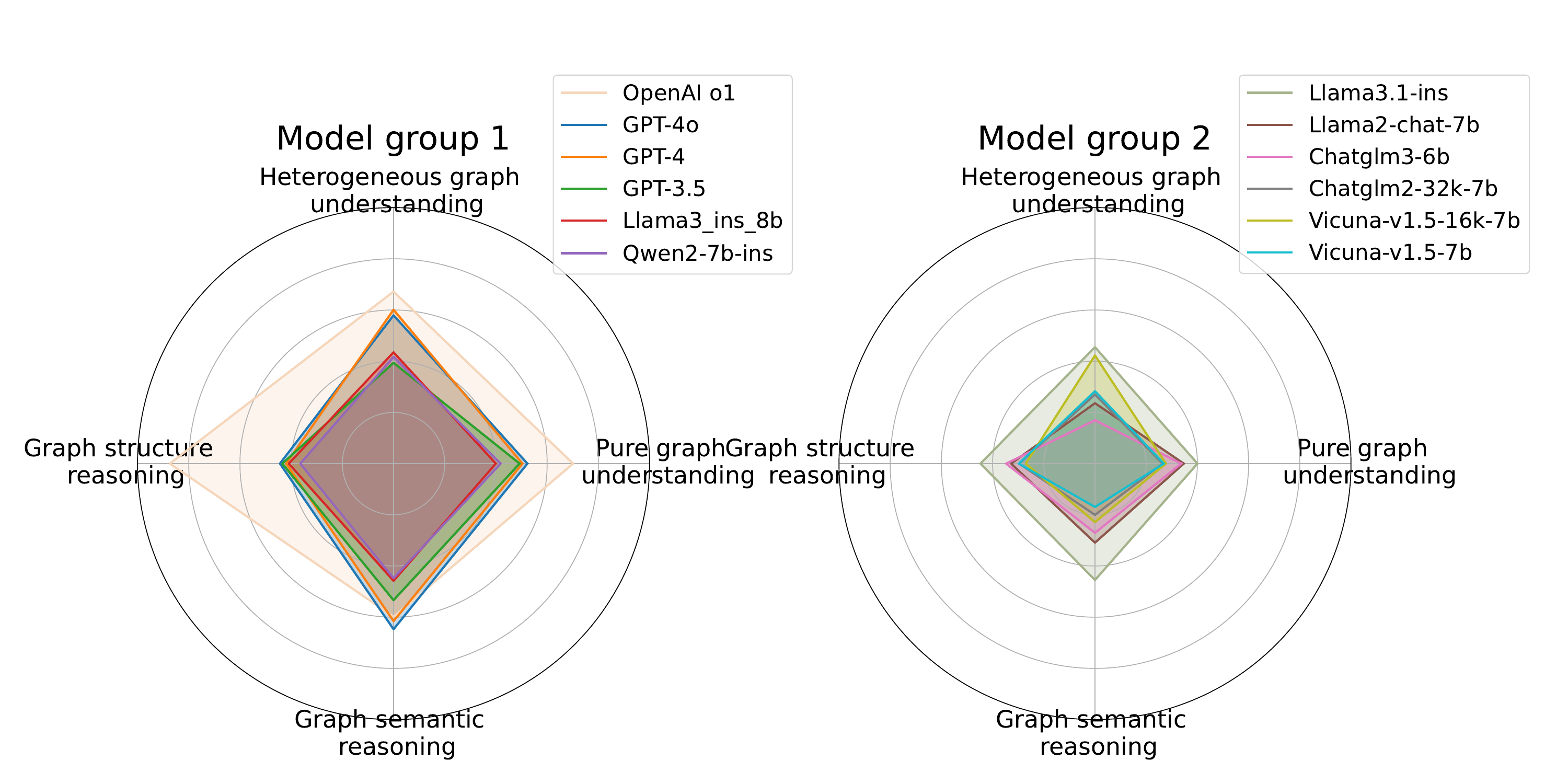} \hfill
  \includegraphics[width=0.5\linewidth]{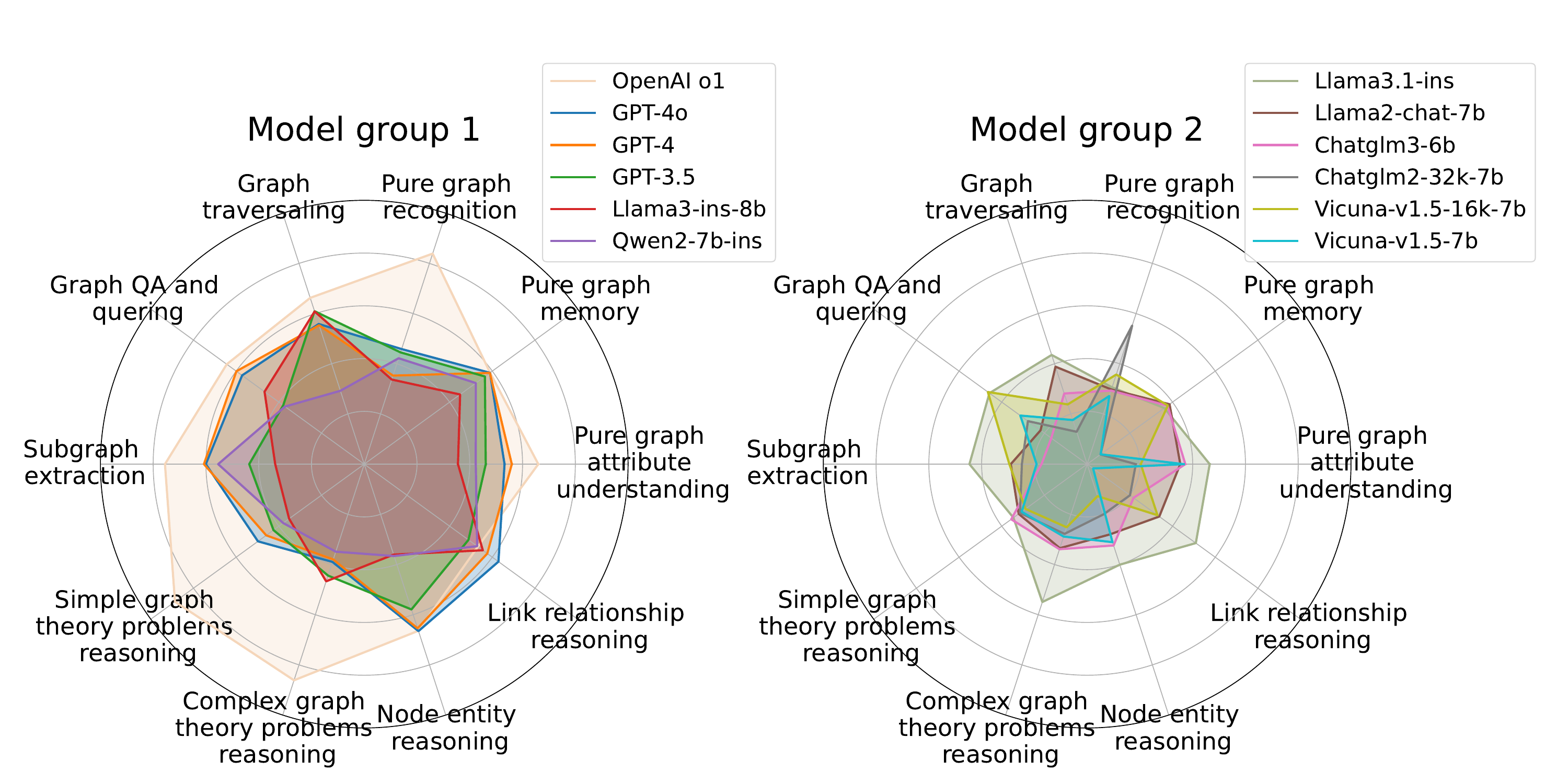}
  \caption {Performance of various LLMs for second and third layer ability dimension.}
    \label{fig:3}
   \vspace{-15pt}
\end{figure*}
\subsection{Evaluation}
This thesis evaluates LLM outputs using exact match accuracy across various output types, including boolean values (e.g., for graph recognition tasks), integers (e.g., path lengths), floating-point numbers (e.g., similarity and average degree), and lists of nodes (e.g., paths). For tasks with multiple valid solutions (e.g., BFS), we check whether the output is a valid solution.

Standardized global scores\citep{standscore}  have become the mainstream choice in fields such as educational assessment and intelligence testing. Compared to using raw scores, employing standardized scores in the multi-task evaluation of LLMs offers the following advantages: 
\begin{itemize}[left=0pt]
    \item  Standardized scores facilitate fair comparisons across different tasks and datasets. Given the varying difficulties and sensitivities of metrics across tasks, absolute performance metrics cannot be directly compared. Standardization normalizes these scores, enabling a more meaningful assessment of model performance across diverse tasks. 
    \item  Standardized scores reduce biases from specific tasks when ranking models. Models often display varying performance across tasks, making it difficult to gauge their strengths in graph understanding and reasoning. By normalizing scores, standardized metrics provide a more accurate and equitable basis for evaluating model capabilities across a broad range of tasks.
\end{itemize}
Since the metrics of different GraCoRe tasks are incomparable and differently sensitive, less experienced audiences cannot easily compare and interpret results, which is also prevalent in recent LLM benchmarks like Kola\citep{kola}.
Therefore, we utilized standardized global scores to evaluate the performance of LLMs on each task. 

Given a task set $T = \{ t_i \}_{i=1}^{|T|}$ and an evaluated model set $M = \{ m_j \}_{j=1}^{|M|}$, so $x_{ij}$ represents the performance of model $m_j$ on task $t_i$. Then the standardized score $z$ can be calculated as:
\begin{equation}
z_{ij} = \frac{x_{ij} - \mu \left( x_{i1}, \ldots, x_{i|M|} \right)}{\sigma \left( x_{i1}, \ldots, x_{i|M|} \right)},
\end{equation}
where $\mu$(·) and $\sigma$(·) denote the mean and standard deviation.Next, we use the Min-Max
scaling\citep{maxmin} method to adjust the scores to the range of 0-100, making it easier to observe and compare the results. The final scores are presented as:
\begin{equation}
s_{ij} = 100 \cdot \frac{z_{ij} - \min(z)}{\max(z) - \min(z)},
\end{equation}
where the functions max ($z$) and min ($z$) correspond to the maximum and minimum of all $z_{ij}$ scores.

\section{Experiments}

\subsection{Experimental Setup}
Using the GraCoRe benchmark, we investigate whether language models can understand and reason about graph structures through textual descriptions. Furthermore, we examine whether tailored prompts can improve their performance on graph-related tasks.

\textbf{Models and Settings} We evaluated a total of eight popular models on the GraCoRe benchmark, including four closed-source models and eight open-source models. The closed-source models are: OpenAI o1 \footnote{This model is an LLMs with powerful reasoning capabilities launched by OpenAI on September 12, 2024. Due to time reasons, we only evaluated it in the main experiment.}, GPT-4o, GPT-4, and GPT-3.5\citep{gpt4}. The open-source models are: LLama3.1-ins-8b, LLama3-ins-8b, LLama2-7b-chat\citep{llama}, Chatglm3-6b, Chatglm2-32k-7b\citep{glm}, Vicuna-v1.5-7b, Vicuna-v1.5-16k-7b\citep{vicuna} and Qwen2-7b-ins\citep{qwen}. More details about these models can be found in the Appendix \ref{sec:B}.

\begin{figure}[t]
  \includegraphics[width=\columnwidth]{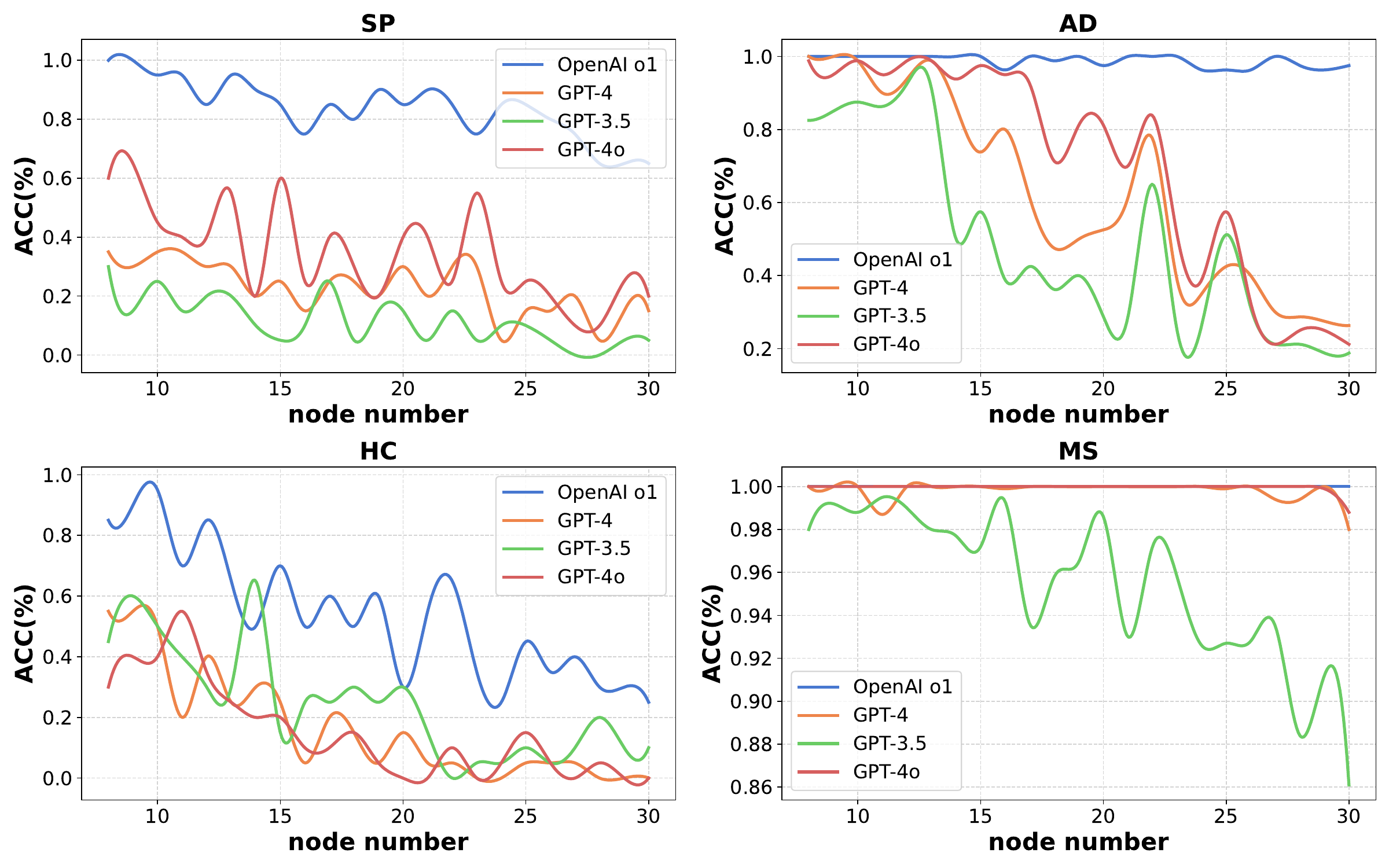}
  \caption{Effect of Graph Size.}
  \label{fig:4}
  \vspace{-15pt}
\end{figure}

\subsection{Main Results}

\textbf{Task Dimensional Analysis} 
Tables \ref{tab:3} and \ref{tab:4} present the performance of various LLMs across 19 tasks in the GraCoRe benchmark. Models generally performed better on graph understanding tasks, while graph reasoning proved more challenging. Closed-source models, particularly OpenAI o1, excelled in both areas, achieving the highest score of 1403.8, significantly surpassing other closed-source models. Among open-source models, Llama3.1-8b and Qwen2-7b-ins also performed well, ranking 5th and 7th with scores of 821.6 and 767.6, respectively. In contrast, Chatglm2-7b underperformed. The overall weaker performance on reasoning tasks was anticipated.

The average $z$-scores reveal a significant gap between open-source and commercial models. Only the Llama3.1-ins-8b and Llama3-ins-8b models achieve $z$-scores above 0, indicating above-average performance. This highlights the need for increased collaboration within the open-source community to enhance large language models. Notably, these two models demonstrate graph understanding and reasoning capabilities approaching those of GPT-3.5.

\noindent\textbf{Ability Dimensional Analysis}
We further analyze Tables \ref{tab:3} and \ref{tab:4} to assess the models' overall performance from a capability perspective, using radar charts to visually represent the second and third layers of the three-tier taxonomy. The left part of Figure \ref{fig:3} illustrates the performance of LLMs across four capability dimensions at the second layer, with each dimension's score averaged across relevant tasks. Most models excel in graph understanding and semantic reasoning but require improvement in structural reasoning. The OpenAI o1 model notably addresses the structural reasoning weaknesses of other OpenAI models, offering a more balanced performance. However, most open-source models underperform in heterogeneous graph understanding tasks.

The right part of Figure \ref{fig:3} provides a detailed analysis of LLM performance across ten capability dimensions at the third layer. It reveals that large language models generally struggle with reasoning in graph theory, especially in complex tasks. However, models like OpenAI o1, GPT-4, and GPT-4o show strong and balanced graph processing abilities, excelling in more complex tasks. Notably, the OpenAI o1 model significantly outperforms others in graph theory reasoning, with performance several times better than other OpenAI models. In contrast, other models display inconsistent performance. For instance, Vicuna-v1.5-16k-7b excels in understanding heterogeneous graph structures but falls behind in other areas, suggesting that rich semantic information may enhance graph processing capabilities in certain contexts.

\noindent\textbf{Long-Text-Specific Models}
Since graph structure data described in text often consists of long texts, the ability of models to handle long texts is also worth noting. As shown in Table 3, models capable of processing long texts, such as Chatglm2-32k-7b and Vicuna-v1.5-16k-7b, performed poorly in graph processing tasks. Compared to other models, their performance was even lower. This suggests that despite being designed for long text input and output, these models still require further development and training to effectively enhance their graph processing capabilities.

Finally, we give the measured values of each model in each task in the main experiment in the Appendix \ref{sec:D}.

\subsection{Further Analysis}

\textbf{Effect of Graph Size }

Figure \ref{fig:4} shows the performance of four OpenAI LLMs on two graph comprehension and two graph reasoning tasks as node count increases. Results reveal a consistent performance decline across all tasks, particularly in the average degree task for graph comprehension, indicating increased computational complexity as node count grows. Despite this, the OpenAI o1 model shows strong computational ability, with minimal impact on comprehension tasks. However, both the shortest path and Hamiltonian path tasks in graph reasoning are notably affected, even for the robust OpenAI o1 and GPT-3.5 models. These findings indicate that current LLMs require further improvement in graph reasoning, particularly with large-scale graphs, where their performance drops considerably.

\noindent\textbf{Effect of Random Sort}
Since our data consists of randomly generated graphs with nodes named by numbers, the performance of node-related tasks may be affected by changes in node order. In this study, we examine the impact of text order on the understanding of graph structures by large language models by comparing random sorting with sequential sorting. The results in Table \ref{table:5} indicate that the model's performance under sequential sorting is often superior to that under random sorting, particularly in graph path reasoning tasks, where the impact is significant. This suggests that renaming nodes and ordering them sequentially can enhance model performance in path reasoning problems. However, it also highlights the model's lack of training on graph data with random sorting. Due to the inference cost, we chose only GPT-3.5 from OpenAI's closed-source models for testing.
\setlength{\tabcolsep}{4pt} 

\begin{table}[t]
\centering
\begin{adjustbox}{max width=\linewidth} 
\begin{tabular}{lcccccc}
\toprule
\textbf{Model} & {}&\textbf{NN} & \textbf{ST} & \textbf{BR} & \textbf{SP} & \textbf{HP} \\
\midrule
\multirow{2}{*}{GPT-3.5} & sort & \textbf{0.993} & \textbf{0.954} & \textbf{0.504} & \textbf{0.117} & \textbf{0.243} \\
 & random & 0.968 & 0.865 & 0.346 & 0.070 & 0.048 \\
\midrule
\multirow{2}{*}{Qwen2-7b} & sort & \textbf{0.876} & 0.874 & \textbf{0.396} & \textbf{0.165} & \textbf{0.115} \\
 & random & 0.773 & \textbf{0.885} & 0.374 & 0.159 & 0.000 \\
\midrule
\multirow{2}{*}{Llama3-ins-8b} & sort & \textbf{0.889} & 0.735 & 0.200 & 0.189 & \textbf{0.285} \\
 & random & 0.786 &\textbf{0.809} & 0.200 & \textbf{0.198} & 0.059 \\
 \midrule
 \multirow{2}{*}{Llama3.1-ins-8b}   & sort        & \textbf{0.706} & \textbf{0.57}  & \textbf{0.217} & \textbf{0.143} & \textbf{0.263} \\
                                   & random sort & 0.52           & 0.552          & 0.198          & 0.102          & 0.243          \\ \midrule
\multirow{2}{*}{Chatglm3-6b}       & sort        & \textbf{0.215} & 0.598          & 0.200  & 0.011 & 0.000    \\
                                   & random sort & 0.196          & \textbf{0.603} & 0.200   & 0.011 & 0.000    \\ \midrule
\multirow{2}{*}{Chatglm2-32k-7b}   & sort        & 0.008          & \textbf{0.005} & \textbf{0.63}  & 0.039          &  0.000    \\
                                   & random sort & \textbf{0.011} & 0.004          & 0.504          & \textbf{0.043} &  0.000   \\ \midrule
\multirow{2}{*}{Llama2-7b-chat-hf} & sort        & 0.075          & \textbf{0.616} & 0.200   & \textbf{0.159} & 0.002 \\
                                   & random sort & \textbf{0.105} & 0.413          & 0.200  & 0.133          & 0.002 \\ 
\bottomrule
\end{tabular}
\end{adjustbox}
\caption{Performance of different models with sorted and random sorted grpah text input.}
\label{table:5}
\vspace{-10pt}
\end{table}

\noindent\textbf{Effect of Text Enhancement}
\begin{figure}[t]
  \includegraphics[width=\columnwidth]{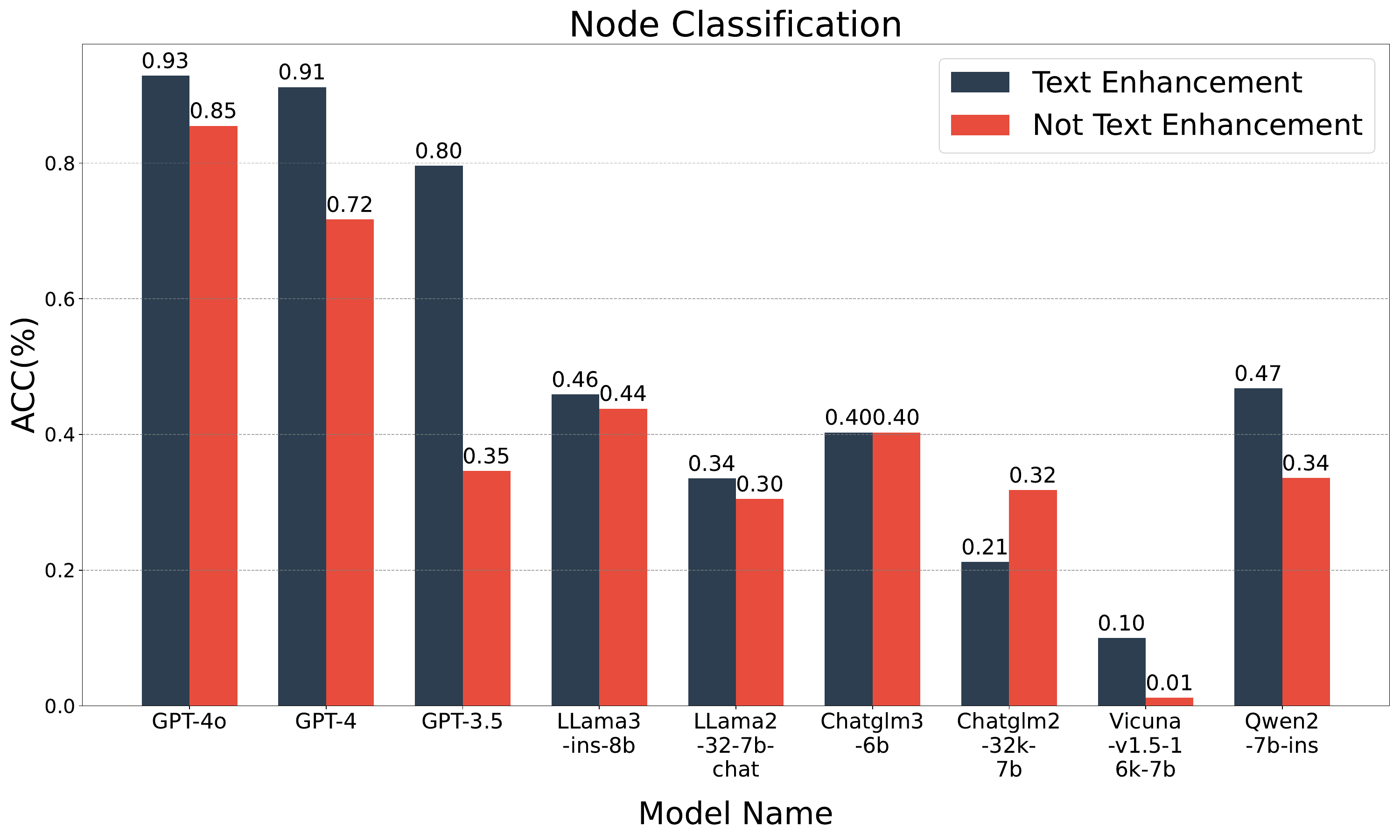}
  \caption{Effect of Text Enhancement.}
  \label{fig:5}
  \vspace{-15pt}
\end{figure}

Heterogeneous graphs provide rich semantic information that aids in understanding and reasoning about graph-related text. To assess whether LLMs can reason purely from structural data, we excluded enhanced text features, such as titles and abstracts, from the ACMText dataset. Figure \ref{fig:5} illustrates the performance of various models in a node classification task. Results show that GPT-4 and GPT-4o maintain strong prediction capabilities without text, indicating effective reasoning based solely on structure. However, GPT-3.5's performance declines significantly without text, while open-source models show minimal impact from its absence.








\section{Conclusion}
This paper introduces GraCoRe, a benchmark designed to evaluate large language models' (LLMs) ability to understand and reason with graph-structured data. We present a detailed, multi-level classification system for assessing model performance on graph-based tasks. Using GraCoRe, we evaluate 12 prominent LLMs, identifying significant limitations in their graph reasoning capabilities. Our experimental results validate the benchmark's effectiveness in measuring LLM performance on graph tasks. Future research will focus on solving complex graph theory problems in large-scale pure graphs, with potential improvements in graph reasoning through methods like agents, chain-of-thought (CoT), and retrieval-augmented generation (RAG).
\section*{Limitations}
As LLMs continue to develop, the volume of training data and their capacity to represent graphs are likely to increase. Our current evaluation may not encompass all their capabilities, and some models might incorporate our data for training, potentially influencing the final evaluation outcomes. In the future, we aim to continually refine and update the GraCoRe benchmark to more effectively assess the graph understanding and reasoning abilities of emerging LLMs.




\section*{Acknowledgments}
The research in this article is supported by  the National Key R\&D Pdrogram of China (Grant No. 2021ZD0112905),  the National Science Foundation of China (Grant No. U22B2059, 62276083),  the Major Key Project of PCL (Grant No. PCL2023A09), the Human-Machine Integrated Consultation System for Cardiovascular Diseases (Grant No. 2023A003). We also appreciate the support from China Mobile Group Heilongjiang Co., Ltd. @ on our research,the research is jointly completed by both parties.

\bibliography{custom}
\clearpage
\appendix
\section{Comparison with existing benchmarks}
\label{sec:A}
In the Introduction, we outlined the advantages of the GraCoRe benchmark. First, it evaluates LLMs' understanding and reasoning abilities on rich graph-structured data while also quantifying their performance on graph-related tasks. Second, GraCoRe tests a wider range of newer models and covers more tasks compared to existing benchmarks. Table \ref{tab:6} highlights the key differences between the GraCoRe benchmark and other benchmarks, demonstrating its broader scope and capabilities.

\section{Details on the data generation and models}
\label{sec:B}
Our dataset consists primarily of two types of graph data: heterogeneous graphs and pure graphs. Based on these, we developed 19 tasks across 11 datasets to evaluate large models. This diversity ensures the comprehensiveness and robustness of our dataset.

\begin{figure*}
  \includegraphics[width=\textwidth]{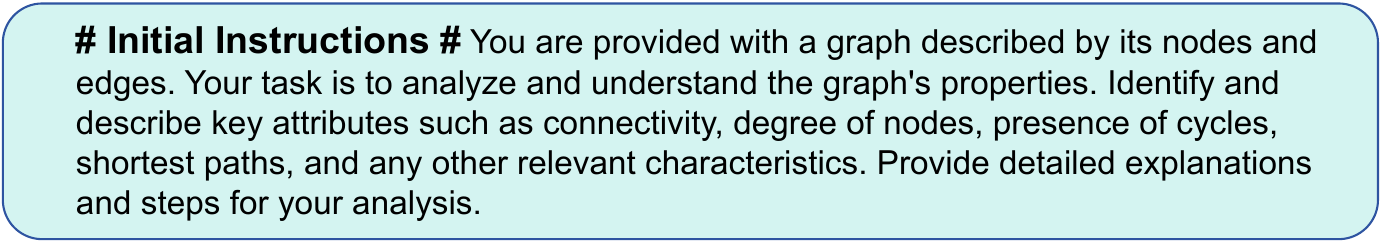}
  \caption{The initial instructions for answer generation.}
  \label{fig:6}
\end{figure*}

\begin{figure*}
  \includegraphics[width=\textwidth]{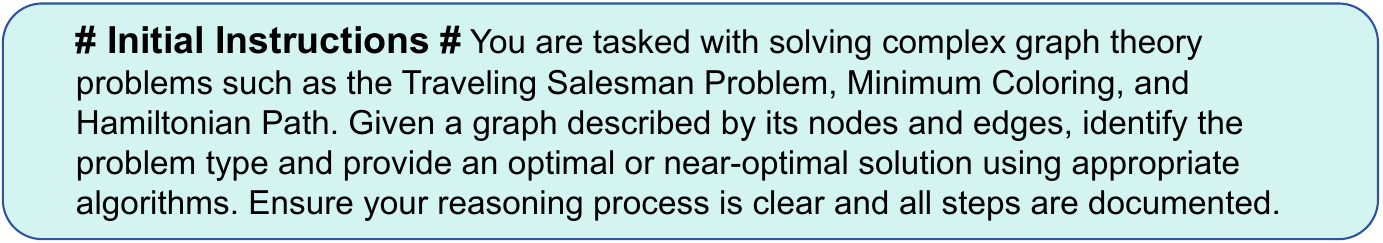}
  \caption{The initial instructions for answer generation.}
  \label{fig:6.1}
\end{figure*}

\begin{figure*}
  \includegraphics[width=\textwidth]{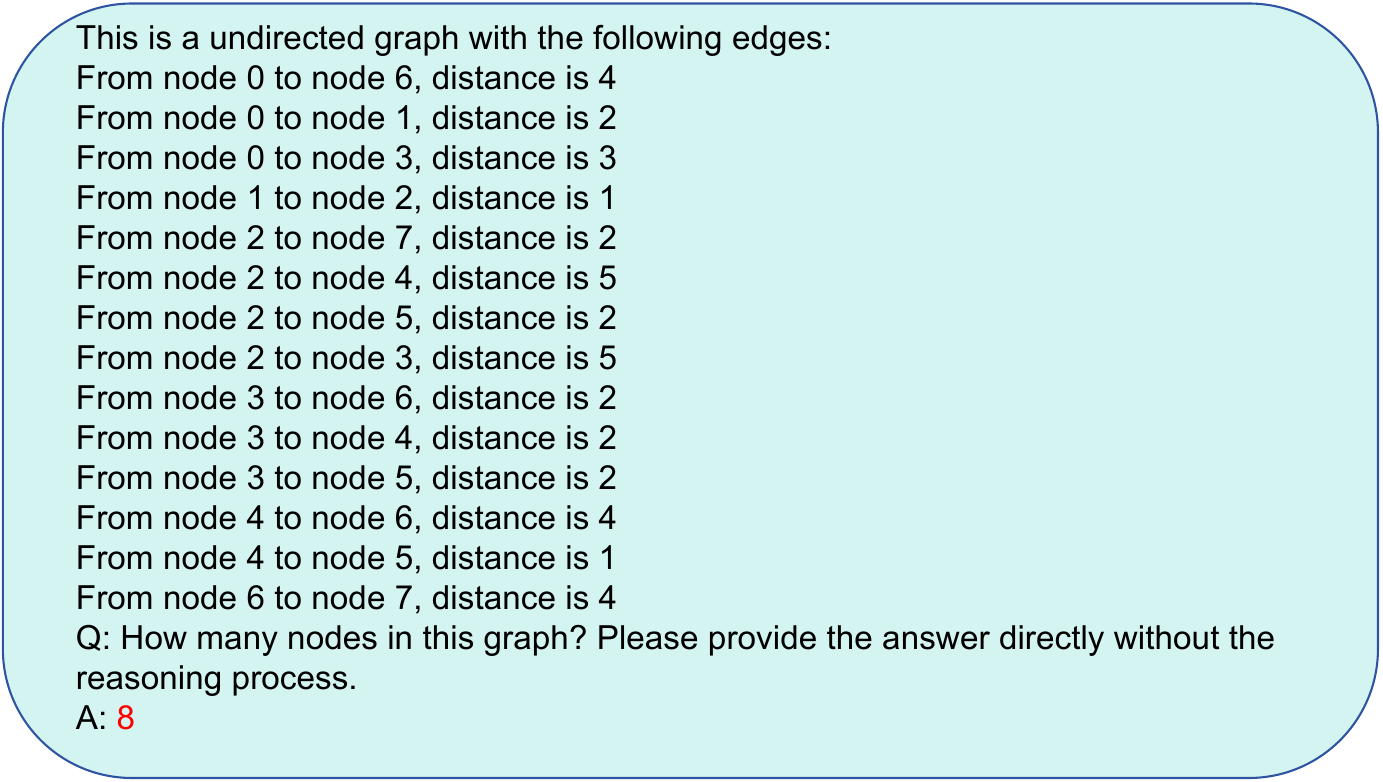}
  \caption{The unique prompt for the Node Number task.}
  \label{fig:7}
\end{figure*}

\begin{figure*}
  \includegraphics[width=\textwidth]{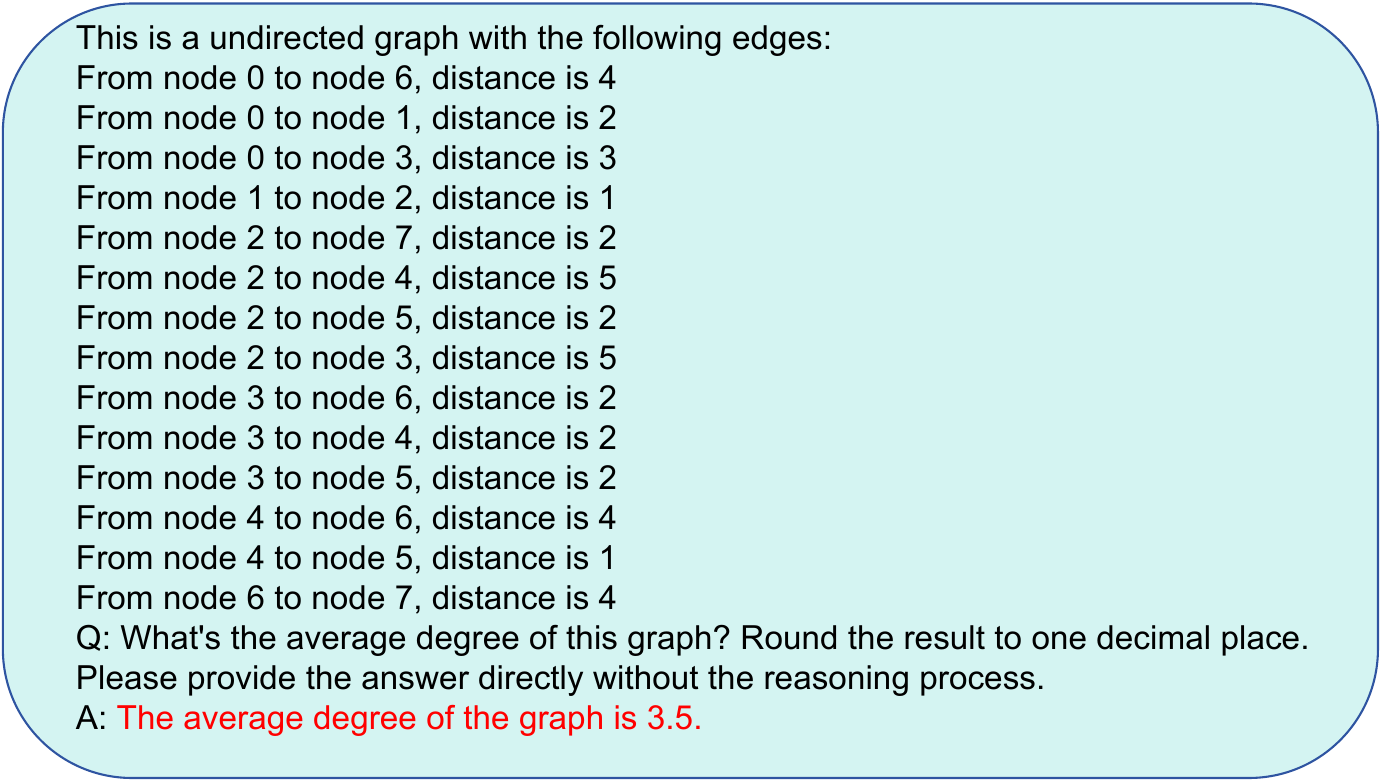}
  \caption{The unique prompt for the  Average Degree task.}
  \label{fig:8}
\end{figure*}

\begin{figure*}
  \includegraphics[width=\textwidth]{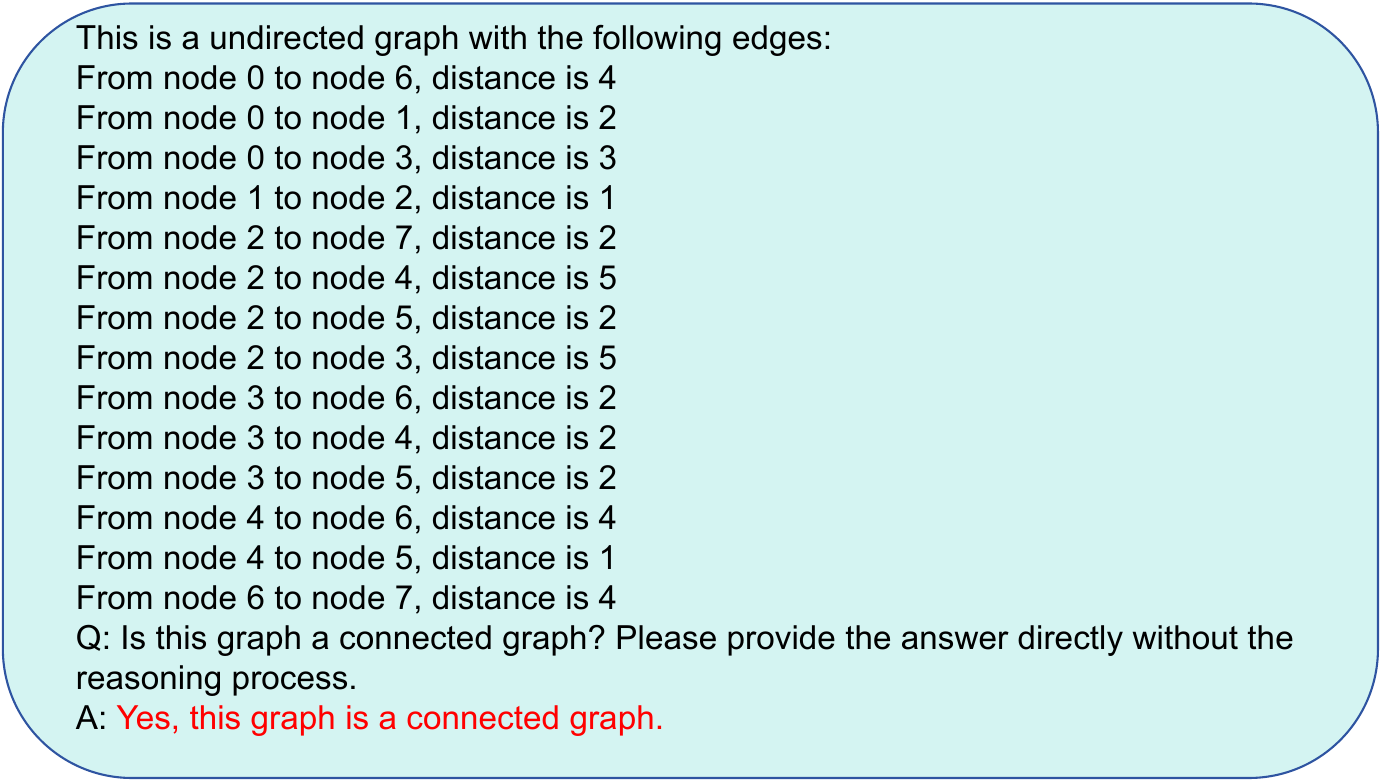}
  \caption{The unique prompt for the Connectivity Test task.}
  \label{fig:9}
\end{figure*}

\begin{figure*}
  \includegraphics[width=\textwidth]{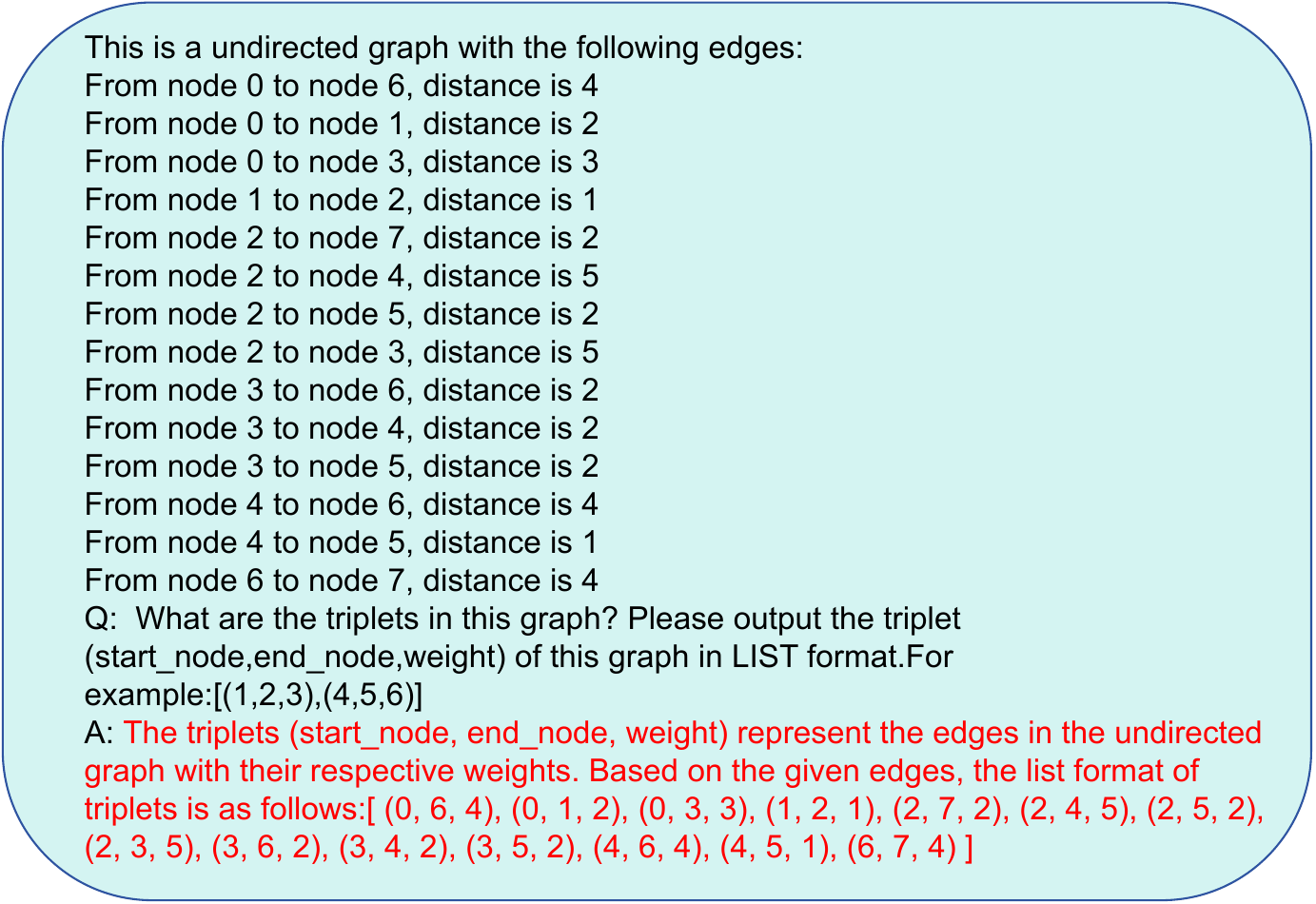}
  \caption{The unique prompt for the Matrix Similarity task.}
  \label{fig:10}
\end{figure*}

\begin{figure*}
  \includegraphics[width=\textwidth]{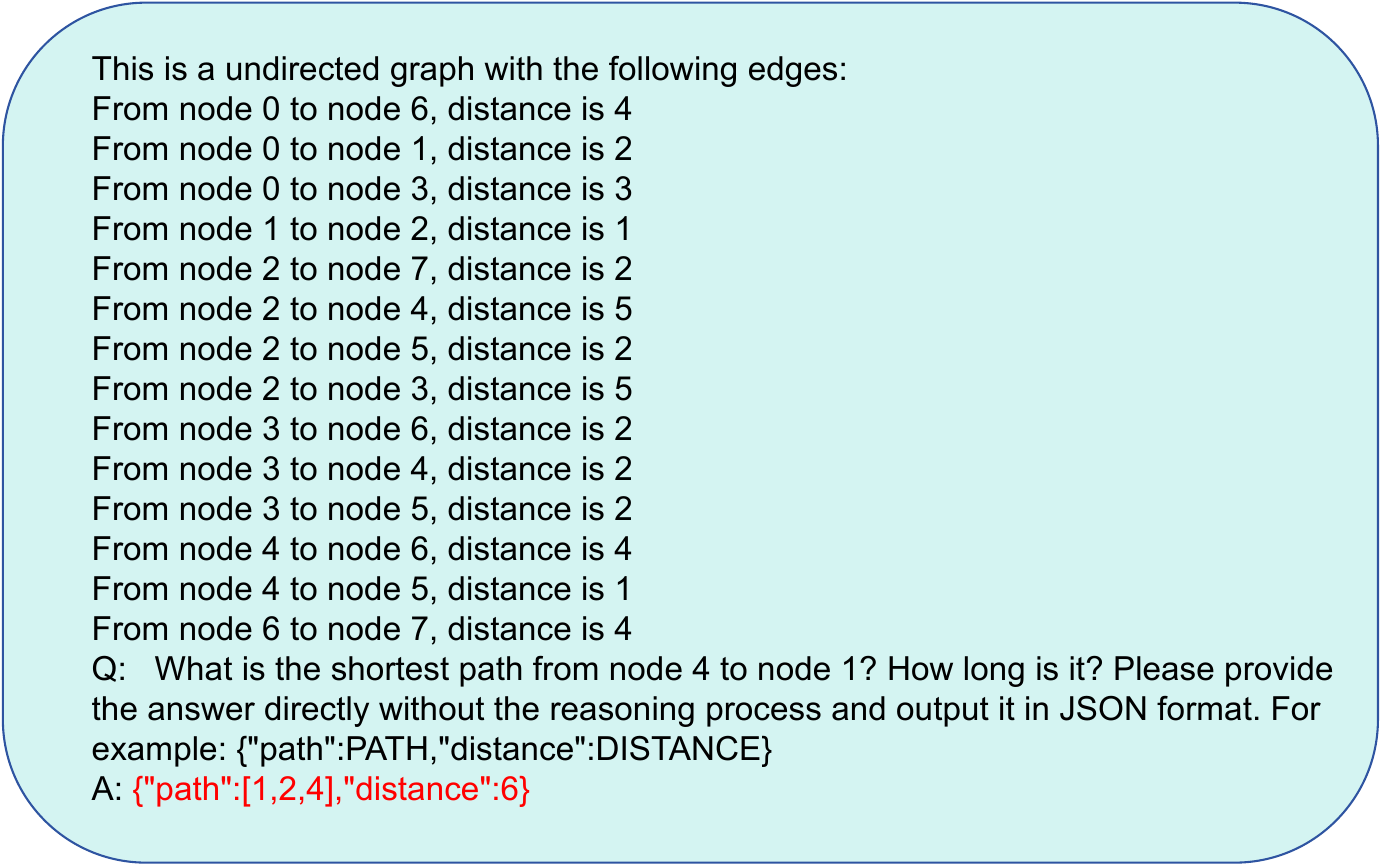}
  \caption{The unique prompt for the  Shortest Path task.}
  \label{fig:11}
\end{figure*}

\begin{figure*}
  \includegraphics[width=\textwidth]{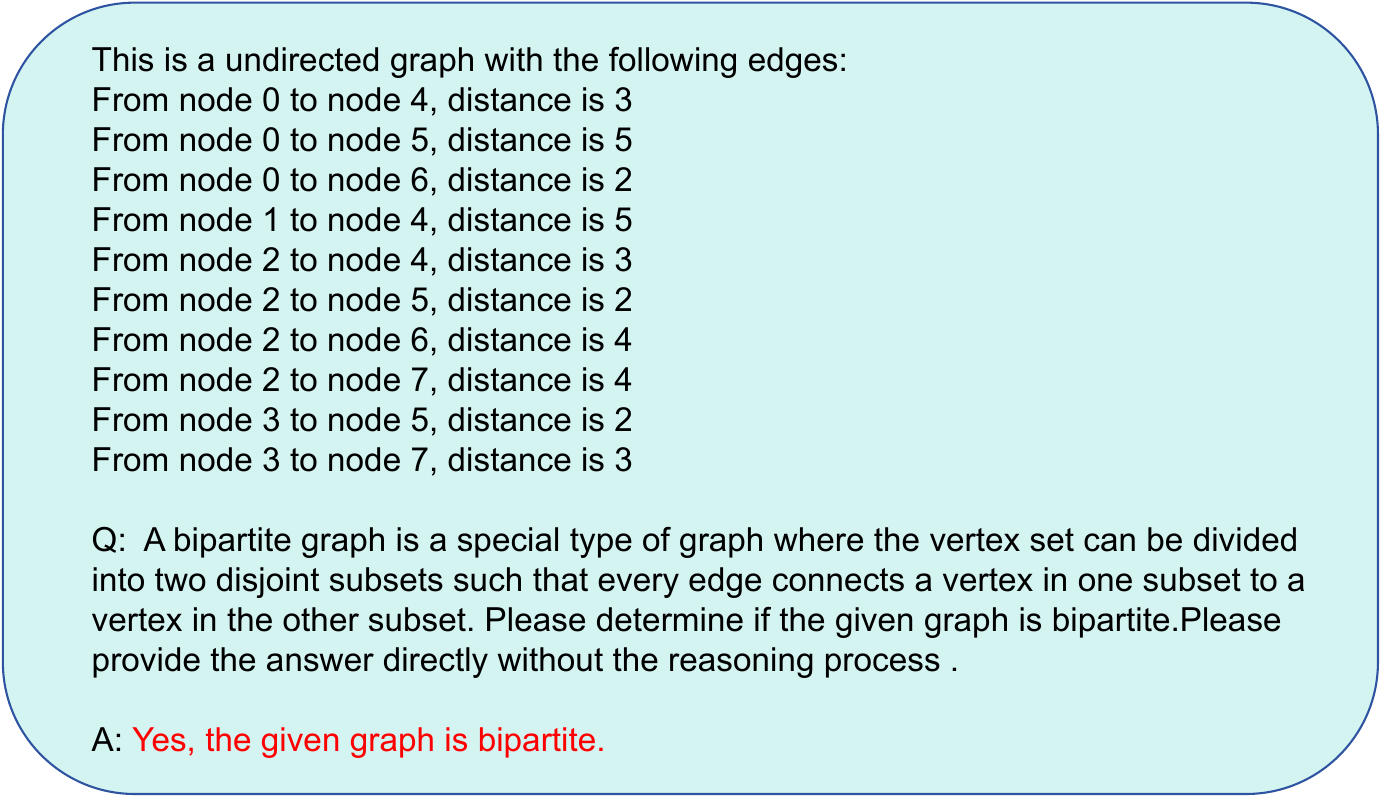}
  \caption{The unique prompt for the Bipartite Recognition task.}
  \label{fig:12}
\end{figure*}

\begin{figure*}
  \includegraphics[width=\textwidth]{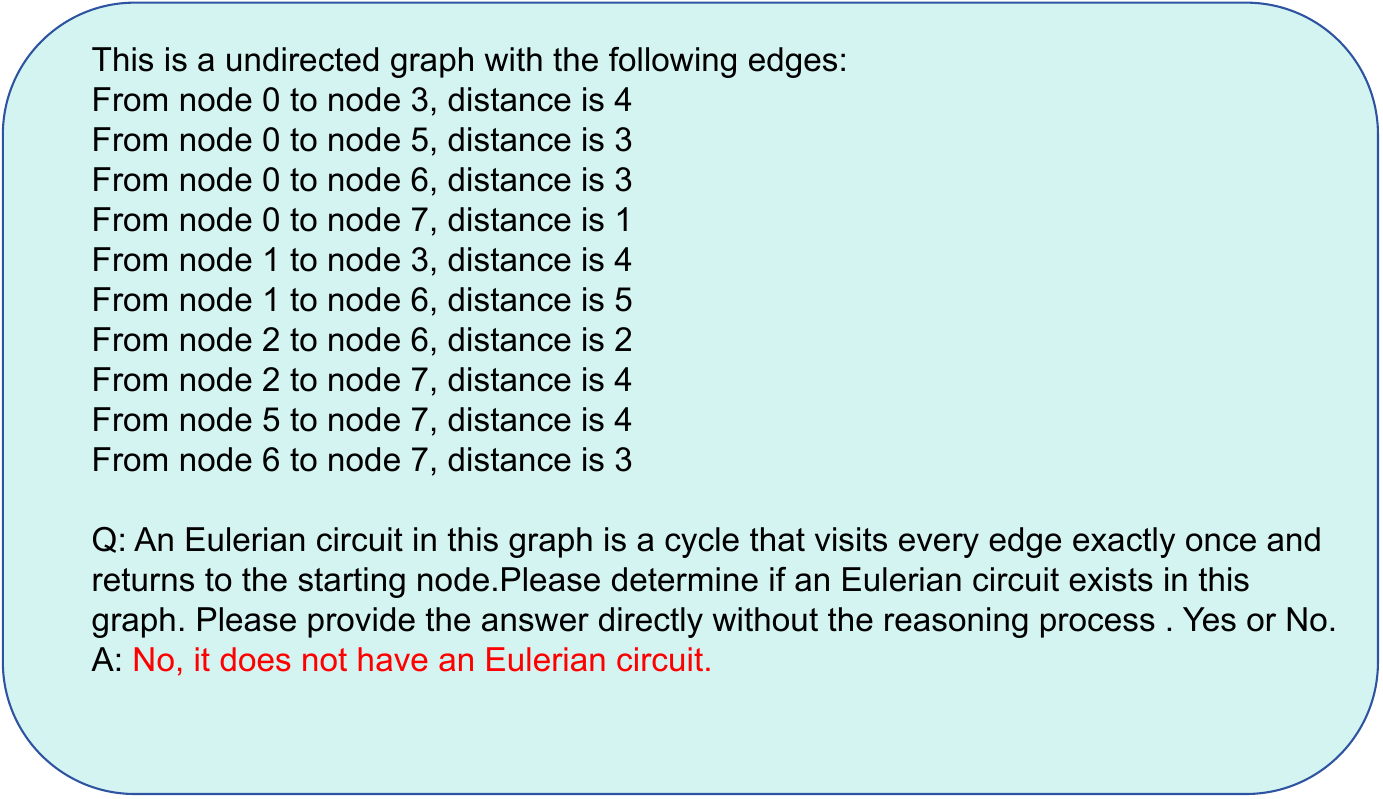}
  \caption{The unique prompt for the Eulerian Path task.}
  \label{fig:13}
\end{figure*}

\begin{figure*}
  \includegraphics[width=\textwidth]{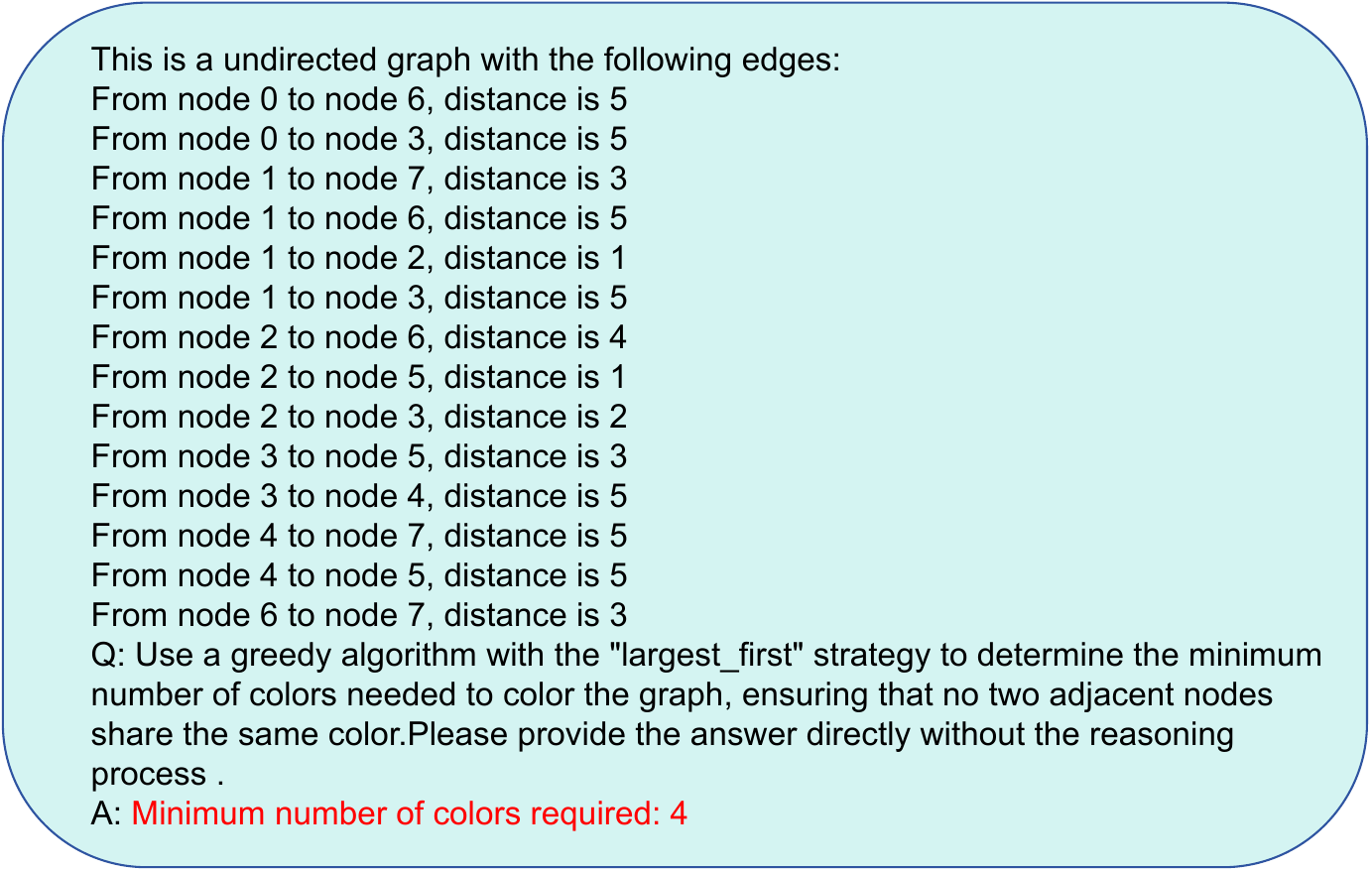}
  \caption{The unique prompt for the Graph Coloring task.}
  \label{fig:14}
\end{figure*}

\begin{figure*}
  \includegraphics[width=\textwidth]{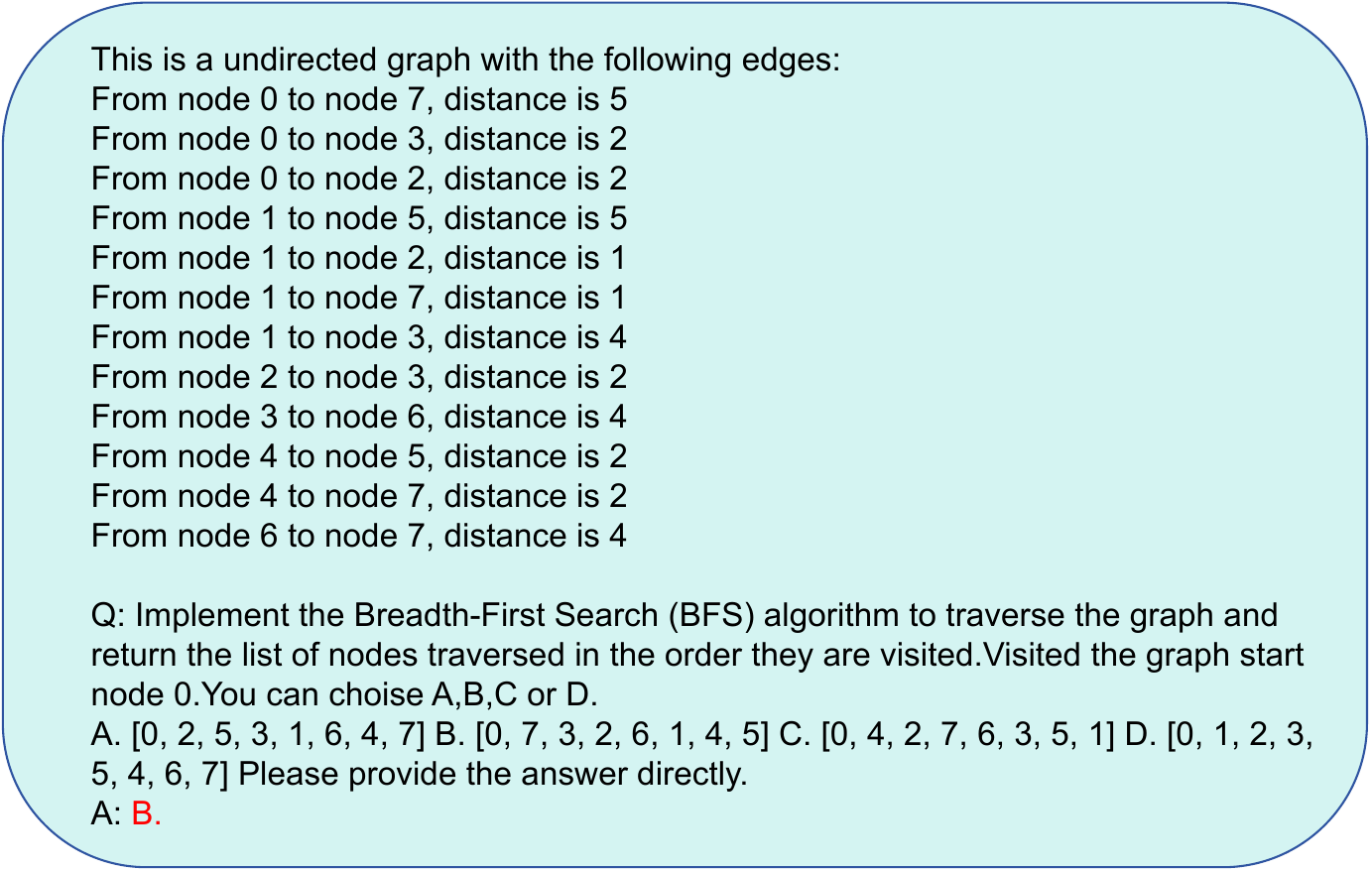}
  \caption{The unique prompt for the Breadth First Search task.}
  \label{fig:15}
\end{figure*}

\begin{figure*}
  \includegraphics[width=\textwidth]{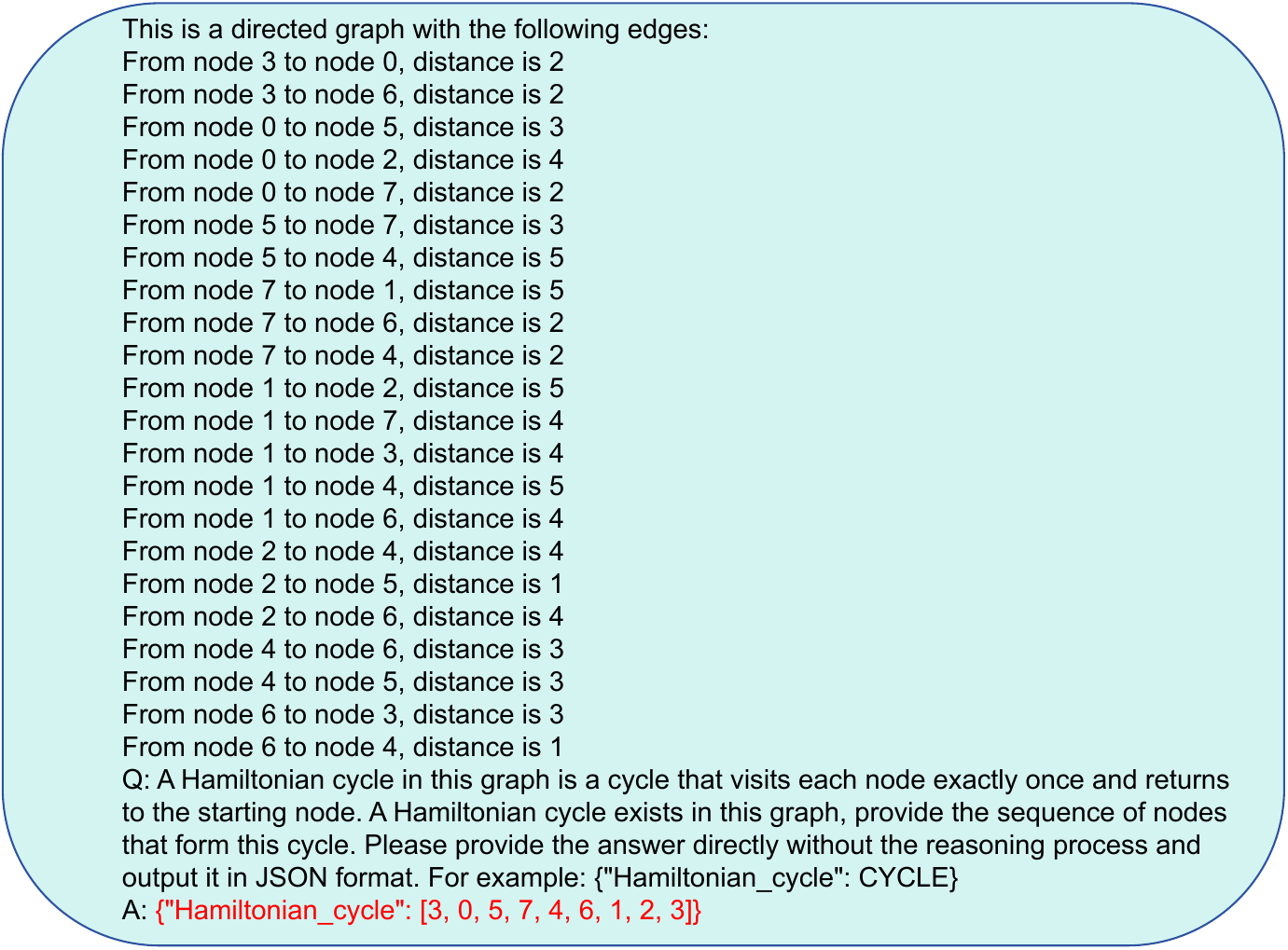}
  \caption{The unique prompt for the  Hamiltonian Cycle task.}
  \label{fig:16}
\end{figure*}

\begin{figure*}
  \includegraphics[width=\textwidth]{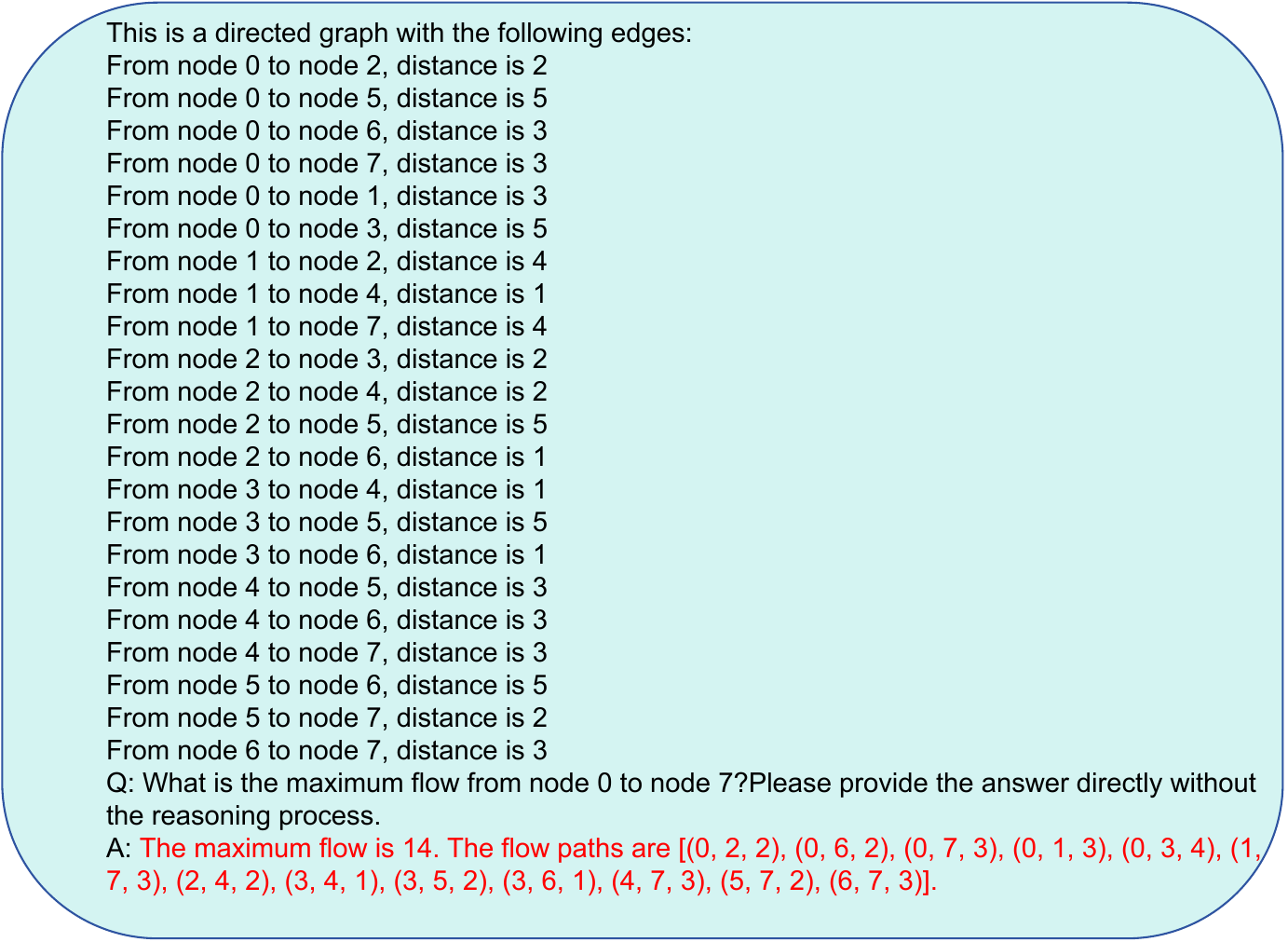}
  \caption{The unique prompt for the Maximum Flow task.}
  \label{fig:17}
\end{figure*}

\begin{figure*}
  \includegraphics[width=\textwidth]{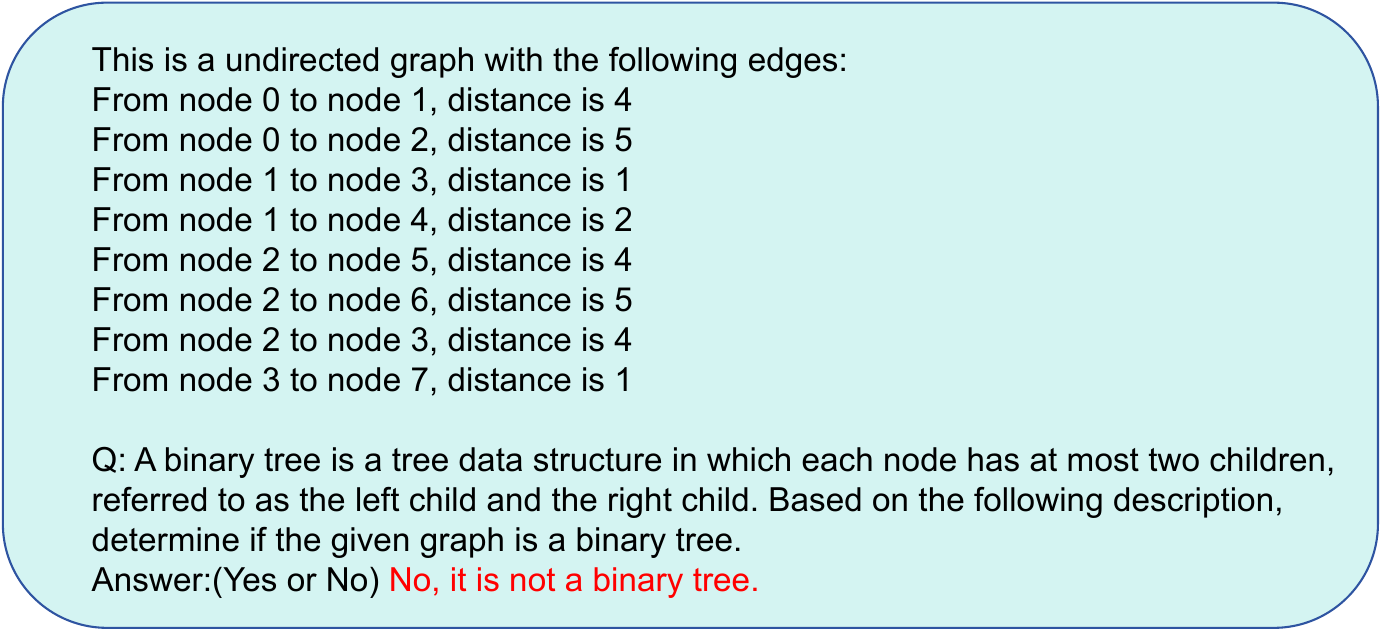}
  \caption{The unique prompt for the Tree Recognition task.}
  \label{fig:18}
\end{figure*}

\begin{figure*}
  \includegraphics[width=\textwidth]{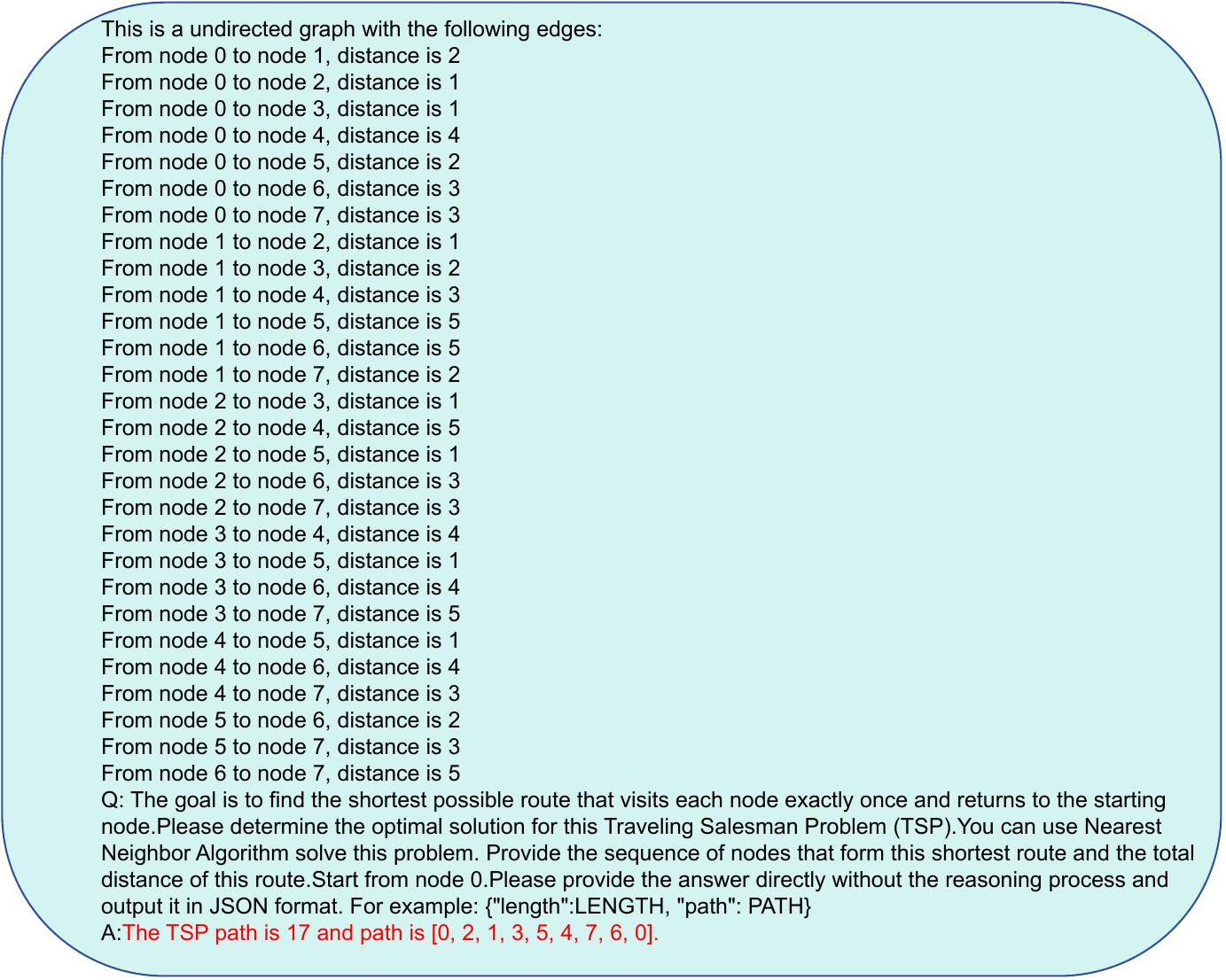}
  \caption{The unique prompt for the  TSP task.}
  \label{fig:19}
\end{figure*}

\begin{figure*}
  \includegraphics[width=\textwidth]{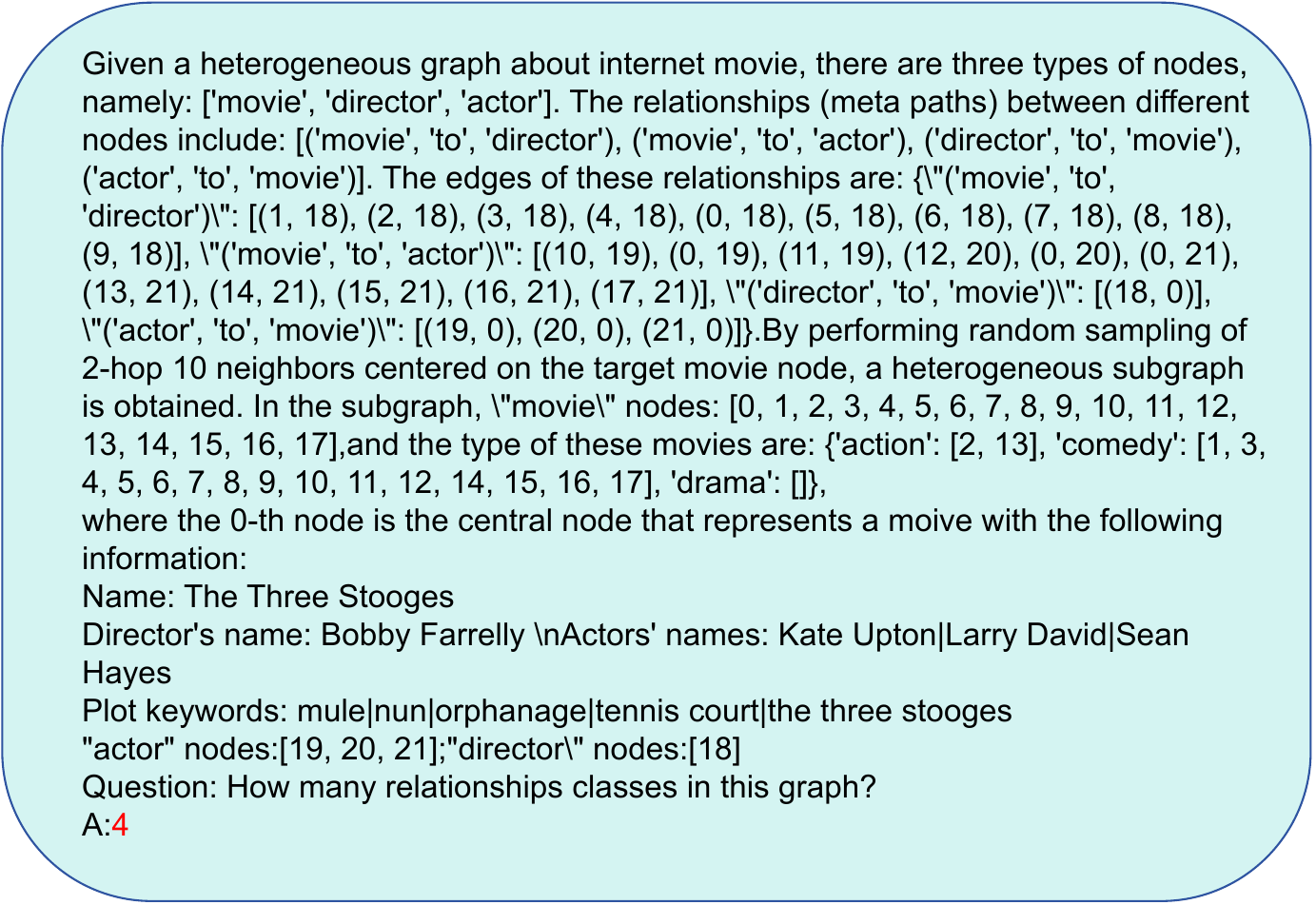}
  \caption{The unique prompt for the  Relation Number task.}
  \label{fig:20}
\end{figure*}

\begin{figure*}
  \includegraphics[width=\textwidth]{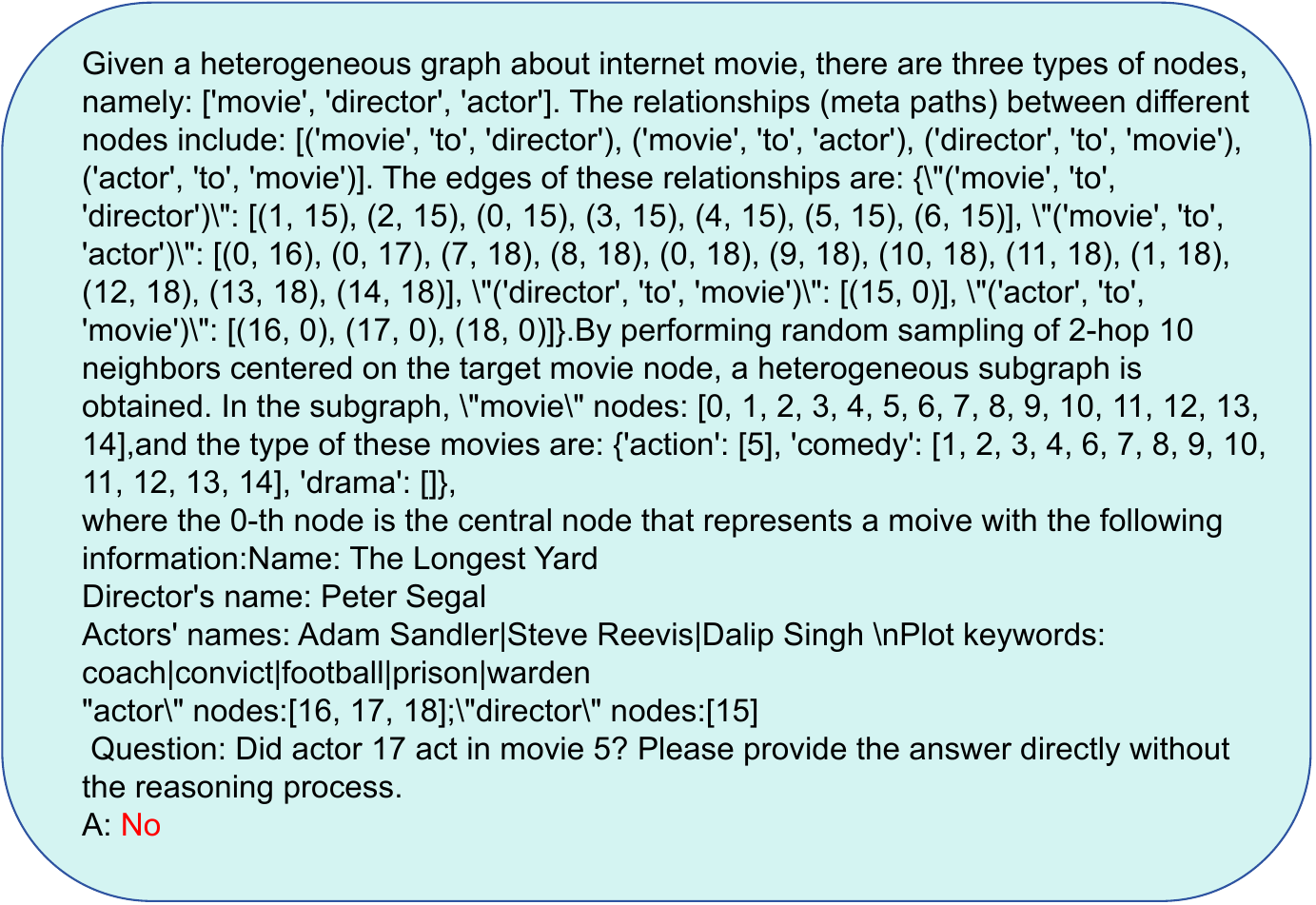}
  \caption{The unique prompt for the Neighborhood Query task.}
  \label{fig:21}
\end{figure*}

\begin{figure*}
  \includegraphics[width=\textwidth]{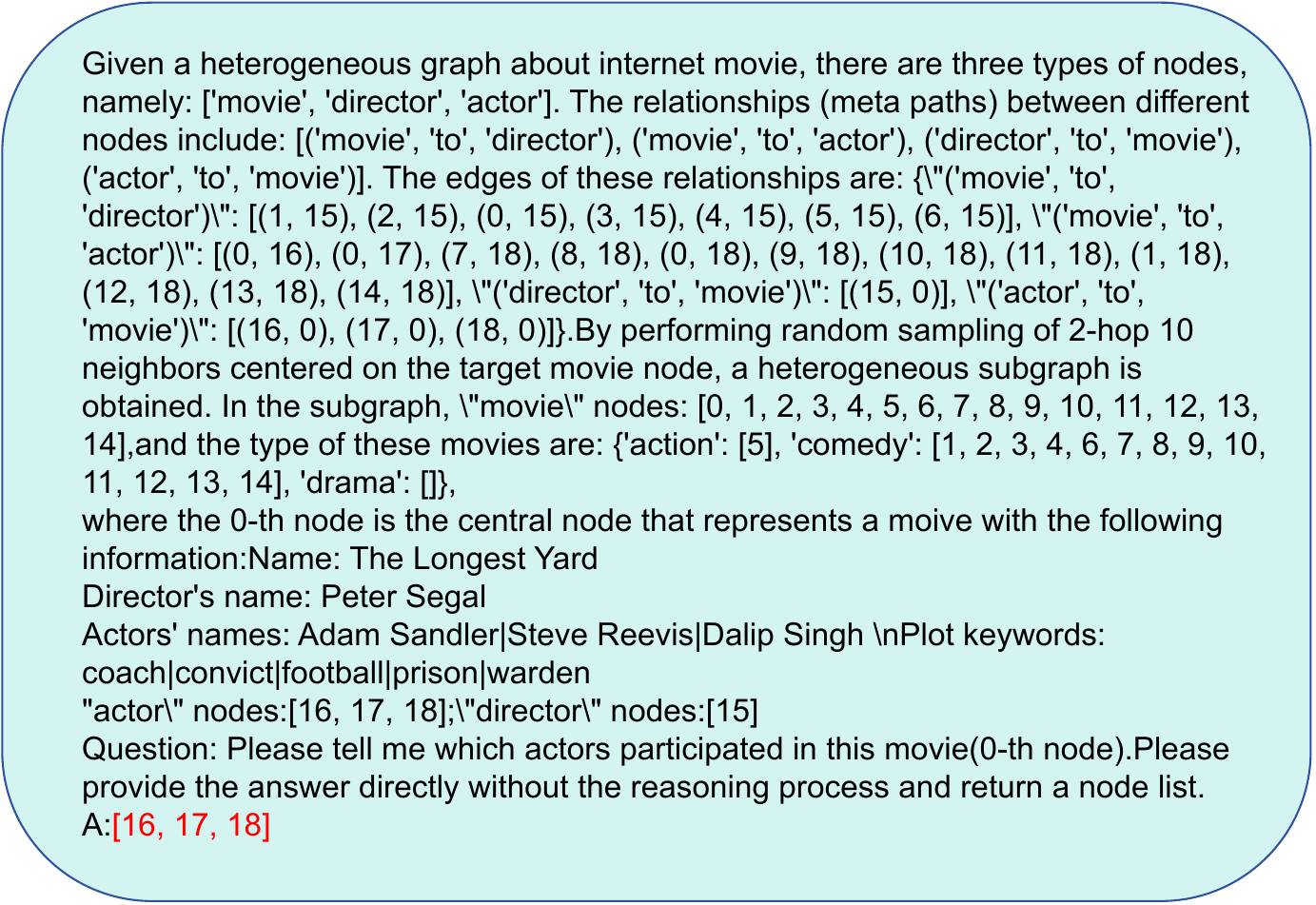}
  \caption{The unique prompt for the Relationship Query task.}
  \label{fig:22}
\end{figure*}

\begin{figure*}
  \includegraphics[width=\textwidth]{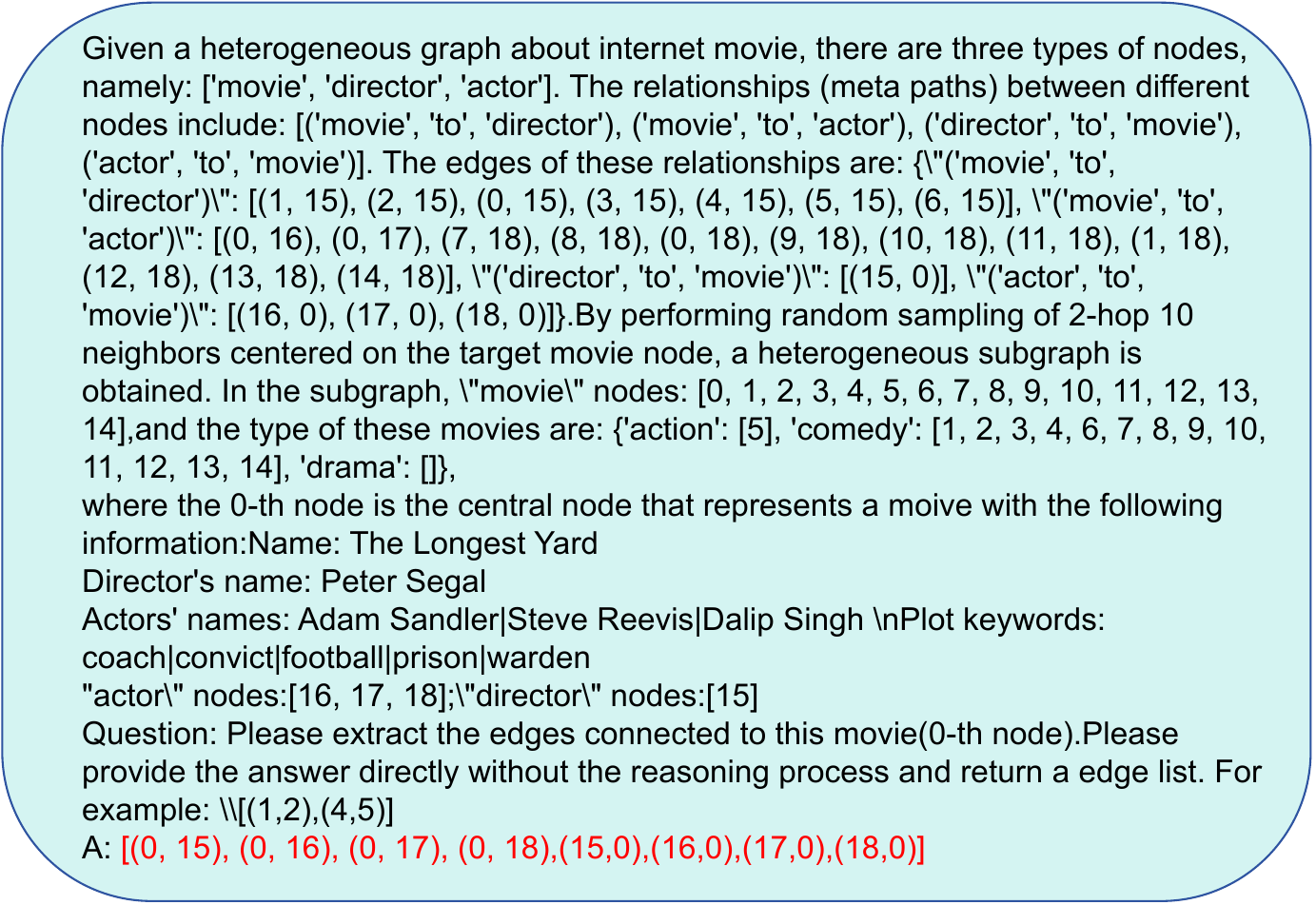}
  \caption{The unique prompt for the  Subgraph Extraction task.}
  \label{fig:23}
\end{figure*}

\begin{figure*}
  \includegraphics[width=\textwidth]{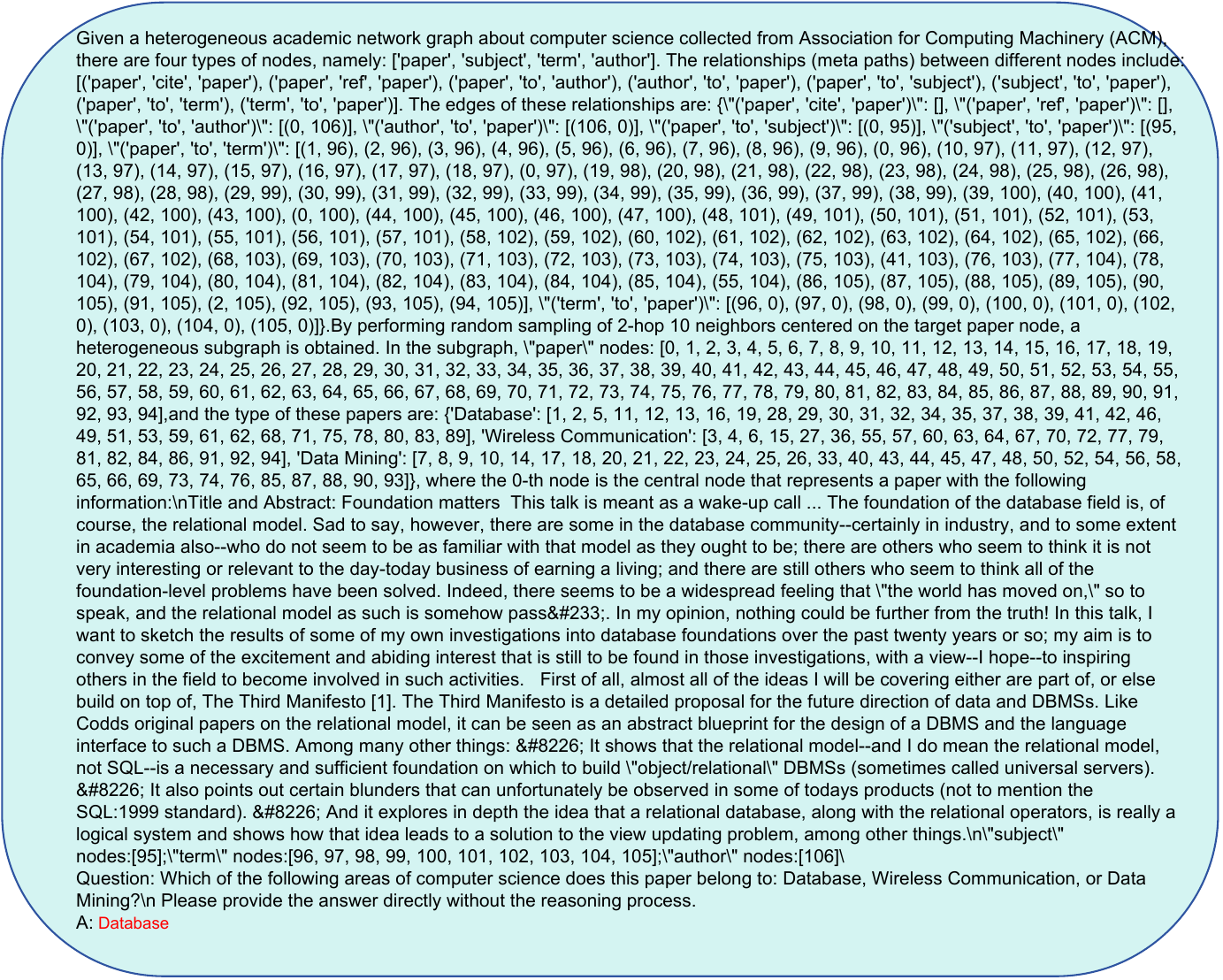}
  \caption{The unique prompt for the Node Classification task.}
  \label{fig:24}
\end{figure*}

\begin{figure*}
  \includegraphics[width=\textwidth]{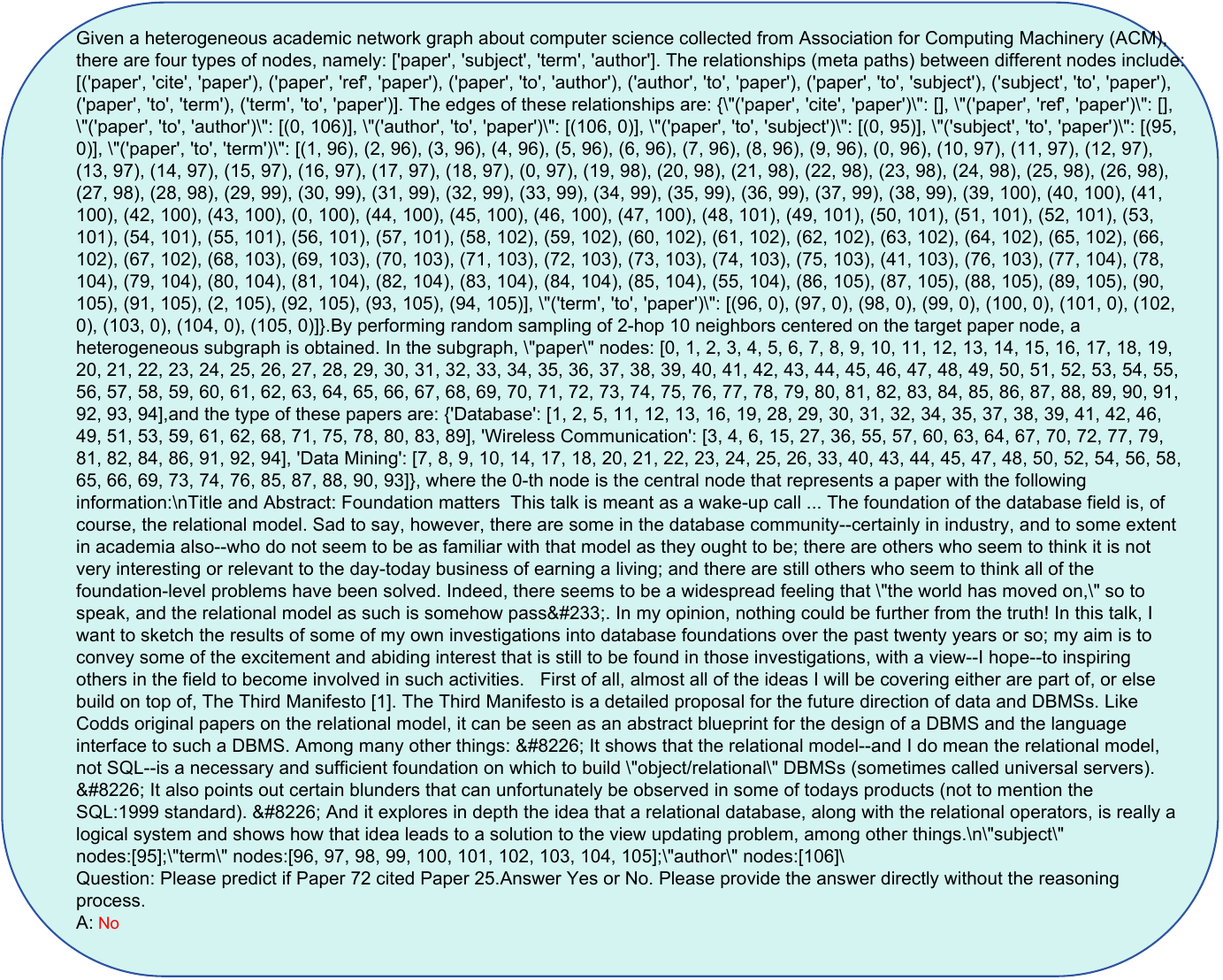}
  \caption{The unique prompt for the Link Prediction task.}
  \label{fig:25}
\end{figure*}

Figures \ref{fig:6} to \ref{fig:25} provide examples of prompts and graph data descriptions for each task. To ensure that generated results meet our capability and task-specific requirements, we concatenate the prompts for each task with the initial prompts.

We evaluated 12 of the latest LLMs, including OpenAI o1 reasoning model, launched on September 12, 2024. Table \ref{tab:7} presents more details on the models and their versions.

\section{More details about each task}
\label{sec:C}

Due to space limitations in the main text, this section will provide a detailed explanation of the specific details for each task. First, we will discuss the classification of model diagram understanding and reasoning capabilities associated with each task, and evaluate the number of test images used for each task. The evaluation results for all tasks are measured using accuracy (ACC). Table \ref{tab:10} summarizes the details of each task.

\section{Raw measurements of each task}
\label{sec:D}
In order to solve the problem of being unable to quantify the capabilities of each model in graph understanding and reasoning tasks, and to compare the differences in capabilities of each model, we use the standardized global scores metric to solve the above problem. However, we will still provide the actual measurement values of the large model in each task, and table \ref{tab:8} and table \ref{tab:9} shows the actual measurement values.

\begin{table*}[htbp]
\centering

\label{tab:benchmark_comparison}
\begin{tabular}{lcccccc}
\toprule
\multirow{2}{*}{\textbf{Benchmark}} & \multicolumn{2}{c}{\textbf{Graph Type}} & \multicolumn{2}{c}{\textbf{Evaluation Perspective}} & \multirow{2}{*}{\textbf{Task \#}} & \multirow{2}{*}{\textbf{Model \#}} \\
\cmidrule(lr){2-3} \cmidrule(lr){4-5}
 & \textbf{Pure} & \textbf{Heterogeneous} & \textbf{Task} & \textbf{Model} & & \\
\midrule
NLGraph\citep{NLgraph-6}       & \CheckmarkBold & \XSolidBold  & \CheckmarkBold & \XSolidBold  & 8  & 2 \\
GPT4Graph\citep{gpt4graph-7}     & \XSolidBold  & \CheckmarkBold & \CheckmarkBold & \XSolidBold  & 10 & 4 \\
GraphArena\citep{tang2024grapharena}    & \XSolidBold  & \CheckmarkBold & \CheckmarkBold & \XSolidBold  & 10 & 10 \\
GraphInstruct\citep{11} & \CheckmarkBold & \XSolidBold  & \CheckmarkBold & \XSolidBold  & 21 & 6 \\
\textbf{GraCoRe}       & \CheckmarkBold & \CheckmarkBold & \CheckmarkBold &\CheckmarkBold & 19 & 12 \\
\bottomrule
\end{tabular}
\caption{Differences between GraCoRe and existing benchmarks.}
\label{tab:6}
\end{table*}

\begin{table*}[htbp]
    \centering
    \small
    \resizebox{\linewidth}{!}{

    \begin{tabular}{ccc}
        \toprule
        \textbf{Model}           & \textbf{Version}           & \textbf{Model Link} \\ 
        \midrule
        \textit{OpenAI o1}       & o1-mini                    & \url{https://platform.openai.com/docs/models/o1} \\
        \textit{GPT-4o}          & gpt-4o                     & \url{https://platform.openai.com/docs/models/gpt-4o} \\
        \textit{GPT-4}           & gpt-4-turbo                & \url{https://platform.openai.com/docs/models/gpt-4-turbo-and-gpt-4} \\
        \textit{GPT-3.5}         & gpt-3.5-turbo              & \url{https://platform.openai.com/docs/models/gpt-3-5-turbo}\\
        \textit{Llama3.1-ins-8b} & Meta-Llama-3.1-8B-Instruct & \url{https://huggingface.co/meta-llama/Meta-Llama-3.1-8B-Instruct} \\
        \textit{Llama3-ins-8b}   & Meta-Llama-3-8B-Instruct   & \url{https://huggingface.co/meta-llama/Meta-Llama-3-8B-Instruct}\\
        \textit{Llama2-7b-chat}  & Llama-2-7b-chat-hf         & \url{https://huggingface.co/meta-llama/Llama-2-7b-chat-hf} \\
        \textit{Chatglm3-6b}     & chatglm3-6b                & \url{https://huggingface.co/THUDM/chatglm3-6b}\\
        \textit{Chatglm2-32k-7b} & chatglm2-6b-32k            & \url{https://huggingface.co/THUDM/chatglm2-6b-32k} \\
        \textit{Vicuna-v1.5-16k} & vicuna-7b-v1.5-16k         & \url{https://huggingface.co/lmsys/vicuna-7b-v1.5-16k} \\
        \textit{Qwen2-7b-ins}    & Qwen2-7B-Instruct          & \url{https://huggingface.co/Qwen/Qwen2-7B-Instruct}\\
        \textit{Vicuna-v1.5-7b}  & vicuna-7b-v1.5             & \url{https://huggingface.co/lmsys/vicuna-7b-v1.5} \\
        \bottomrule
    \end{tabular}}
    \caption{More details about models.}
    \label{tab:7}
\end{table*}

\begin{table*}[htbp]
	\centering
	\scriptsize
	\setlength{\extrarowheight}{2.5pt} 
        \resizebox{\linewidth}{!}{
	\begin{tabular}{c|ccccccc|cccc}
		\hline
		\multirow{3}{*}{\textbf{Model}} & \multicolumn{7}{c|}{\textbf{Pure Graph}}                                                                                                                                                                                                                                                                                                                                        & \multicolumn{4}{c}{\textbf{Heterogeneous Graph}}                                                                                                                                             \\ \cline{2-12} 
		& \multicolumn{3}{c|}{\textbf{\begin{tabular}[c]{@{}c@{}}Attribute\\  Understanding\end{tabular}}}                   & \multicolumn{1}{c|}{\textbf{\begin{tabular}[c]{@{}c@{}}Graph \\ Memory\end{tabular}}} & \multicolumn{2}{c|}{\textbf{\begin{tabular}[c]{@{}c@{}}Graph \\ Recognition\end{tabular}}} & \textbf{\begin{tabular}[c]{@{}c@{}}Grap \\ Traversaling\end{tabular}} & \multicolumn{3}{c|}{\textbf{\begin{tabular}[c]{@{}c@{}}Graph QA \\ and Quering\end{tabular}}}                      & \textbf{\begin{tabular}[c]{@{}c@{}}Subgraph \\ Extraction\end{tabular}} \\ \cline{2-12} 
		& \multicolumn{1}{c|}{\textbf{NN(ACC)}}     & \multicolumn{1}{c|}{\textbf{AD(ACC)}}     & \multicolumn{1}{c|}{\textbf{CT(ACC)}}     & \multicolumn{1}{c|}{\textbf{MS(ACC)}}                                                      & \multicolumn{1}{c|}{\textbf{TR(ACC)}}              & \multicolumn{1}{c|}{\textbf{BR(ACC)}}           & \textbf{BFS(ACC)}                                                          & \multicolumn{1}{c|}{\textbf{NQ(ACC)}}     & \multicolumn{1}{c|}{\textbf{RQ(ACC)}}     & \multicolumn{1}{c|}{\textbf{RN(ACC)}}     & \textbf{SE(ACC)}                                                             \\ \hline
\textit{o1-mini}         & \multicolumn{1}{c|}{0.968}          & \multicolumn{1}{c|}{\textbf{0.988}}   & \multicolumn{1}{c|}{\textbf{0.978}}                   & \multicolumn{1}{c|}{\textbf{1.000}}    & \multicolumn{1}{c|}{\textbf{0.998}}               & \multicolumn{1}{c|}{\textbf{0.600}}                   & \multicolumn{1}{c|}{\textbf{0.991}}     & \multicolumn{1}{c|}{\textbf{0.978}}   & \multicolumn{1}{c|}{\textbf{0.619}}     & \multicolumn{1}{c|}{0.986}           & \textbf{0.971}      \\
\textit{GPT-4o}          & \multicolumn{1}{c|}{\textbf{0.993}} & \multicolumn{1}{c|}{0.706}            & \multicolumn{1}{c|}{0.753}                            & \multicolumn{1}{c|}{0.999}             & \multicolumn{1}{c|}{0.630}                        & \multicolumn{1}{c|}{0.339}                            & \multicolumn{1}{c|}{0.809}              & \multicolumn{1}{c|}{0.745}            & \multicolumn{1}{c|}{0.562}              & \multicolumn{1}{c|}{0.955}           & 0.713               \\
\textit{GPT-4}           & \multicolumn{1}{c|}{0.991}          & \multicolumn{1}{c|}{0.626}            & \multicolumn{1}{c|}{0.946}                            & \multicolumn{1}{c|}{0.998}             & \multicolumn{1}{c|}{0.578}                        & \multicolumn{1}{c|}{0.211}                            & \multicolumn{1}{c|}{0.809}              & \multicolumn{1}{c|}{0.843}            & \multicolumn{1}{c|}{0.560}              & \multicolumn{1}{c|}{\textbf{1.000}}  & 0.725               \\
\textit{GPT-3.5}         & \multicolumn{1}{c|}{\textbf{0.993}} & \multicolumn{1}{c|}{0.490}            & \multicolumn{1}{c|}{0.698}                            & \multicolumn{1}{c|}{0.954}             & \multicolumn{1}{c|}{{\color[HTML]{1F2329} 0.459}} & \multicolumn{1}{c|}{0.504}                            & \multicolumn{1}{c|}{0.898}              & \multicolumn{1}{c|}{0.489}            & \multicolumn{1}{c|}{0.291}              & \multicolumn{1}{c|}{0.637}           & 0.439               \\
\textit{Llama3.1-ins-8b} & \multicolumn{1}{c|}{0.706}          & \multicolumn{1}{c|}{0.689}            & \multicolumn{1}{c|}{0.678}                            & \multicolumn{1}{c|}{0.570}             & \multicolumn{1}{c|}{0.498}                        & \multicolumn{1}{c|}{0.217}                            & \multicolumn{1}{c|}{0.593}              & \multicolumn{1}{c|}{0.670}            & \multicolumn{1}{c|}{0.534}              & \multicolumn{1}{c|}{0.389}           & 0.456               \\
\textit{Llama3-ins-8b}   & \multicolumn{1}{c|}{0.889}          & \multicolumn{1}{c|}{0.289}            & \multicolumn{1}{c|}{0.555}                            & \multicolumn{1}{c|}{0.735}             & \multicolumn{1}{c|}{0.563}                        & \multicolumn{1}{c|}{0.200}                            & \multicolumn{1}{c|}{0.896}              & \multicolumn{1}{c|}{0.295}            & \multicolumn{1}{c|}{0.556}              & \multicolumn{1}{c|}{0.892}           & 0.275               \\
\textit{Llama2-7b-chat}  & \multicolumn{1}{c|}{0.075}          & \multicolumn{1}{c|}{0.411}            & \multicolumn{1}{c|}{0.907}                            & \multicolumn{1}{c|}{0.616}             & \multicolumn{1}{c|}{0.500}                        & \multicolumn{1}{c|}{0.200}                            & \multicolumn{1}{c|}{0.511}              & \multicolumn{1}{c|}{0.285}            & \multicolumn{1}{c|}{0.035}              & \multicolumn{1}{c|}{0.418}           & 0.195               \\
\textit{Chatglm3-6b}     & \multicolumn{1}{c|}{0.215}          & \multicolumn{1}{c|}{0.377}            & \multicolumn{1}{c|}{0.934}                            & \multicolumn{1}{c|}{0.598}             & \multicolumn{1}{c|}{0.493}                        & \multicolumn{1}{c|}{0.200}                            & \multicolumn{1}{c|}{0.324}              & \multicolumn{1}{c|}{0.039}            & \multicolumn{1}{c|}{0.238}              & \multicolumn{1}{c|}{0.000}           & 0.001               \\
\textit{Chatglm2-32k-7b} & \multicolumn{1}{c|}{0.008}          & \multicolumn{1}{c|}{0.255}            & \multicolumn{1}{c|}{0.446}                            & \multicolumn{1}{c|}{0.005}             & \multicolumn{1}{c|}{0.515}                        & \multicolumn{1}{c|}{0.630}                            & \multicolumn{1}{c|}{0.057}              & \multicolumn{1}{c|}{0.568}            & \multicolumn{1}{c|}{0.238}              & \multicolumn{1}{c|}{0.000}           & 0.125               \\
\textit{Vicuna-v1.5-16k} & \multicolumn{1}{c|}{0.397}          & \multicolumn{1}{c|}{0.328}            & \multicolumn{1}{c|}{0.210}                            & \multicolumn{1}{c|}{0.601}             & \multicolumn{1}{c|}{0.472}                        & \multicolumn{1}{c|}{0.335}                            & \multicolumn{1}{c|}{0.248}              & \multicolumn{1}{c|}{0.377}            & \multicolumn{1}{c|}{0.542}              & \multicolumn{1}{c|}{0.802}           & 0.204               \\
\textit{Qwen2-7b-ins}    & \multicolumn{1}{c|}{0.876}          & \multicolumn{1}{c|}{0.165}            & \multicolumn{1}{c|}{0.958}                            & \multicolumn{1}{c|}{0.874}             & \multicolumn{1}{c|}{0.520}                        & \multicolumn{1}{c|}{0.396}                            & \multicolumn{1}{c|}{0.343}              & \multicolumn{1}{c|}{0.454}            & \multicolumn{1}{c|}{0.556}              & \multicolumn{1}{c|}{0.079}           & 0.635               \\
\textit{Vicuna-v1.5-7b}  & \multicolumn{1}{c|}{0.110}          & \multicolumn{1}{c|}{0.415}            & \multicolumn{1}{c|}{0.934}                            & \multicolumn{1}{c|}{0.007}             & \multicolumn{1}{c|}{0.459}                        & \multicolumn{1}{c|}{0.200}                            & \multicolumn{1}{c|}{0.139}              & \multicolumn{1}{c|}{0.016}            & \multicolumn{1}{c|}{0.558}              & \multicolumn{1}{c|}{0.281}           & 0.032              
\\ \hline
	\end{tabular}}
     \caption{Performance of graph understanding.}
    \label{tab:8}
\end{table*}

\begin{table*}[htbp]
	\centering
	\scriptsize
	\setlength{\extrarowheight}{3pt} 
	
	\begin{tabular}{c|cccccc|cc}
		\hline
		\multirow{3}{*}{\textbf{Model}} & \multicolumn{6}{c|}{\textbf{Graph Structure Reasoning}}                                                                                                                                                                     & \multicolumn{2}{c}{\textbf{Graph Semantic Reasoning}}                                                                                                                            \\ \cline{2-9}
		& \multicolumn{3}{c|}{\textbf{\begin{tabular}[c]{@{}c@{}}Simple Graph \\ Theory Problems\end{tabular}}}              & \multicolumn{3}{c|}{\textbf{\begin{tabular}[c]{@{}c@{}}Complex Graph \\ Theory Problems\end{tabular}}} & \multicolumn{1}{c|}{\textbf{\begin{tabular}[c]{@{}c@{}}Node Entity \\ Reasoning\end{tabular}}} & \textbf{\begin{tabular}[c]{@{}c@{}}Link Relationship \\ Reasoning\end{tabular}}\\ \cline{2-9}
		& \multicolumn{1}{c|}{\textbf{SP(ACC)}}     & \multicolumn{1}{c|}{\textbf{MF(ACC)}}     & \multicolumn{1}{c|}{\textbf{EP(ACC)}}     & \multicolumn{1}{c|}{\textbf{HC(ACC)}}        & \multicolumn{1}{c|}{\textbf{TSP(ACC)}}       & \textbf{GC(ACC)}        & \multicolumn{1}{c|}{\textbf{NC(ACC)}}                                                               & \textbf{LP(ACC)}  \\ \hline
	\textit{o1-mini}         & \multicolumn{1}{c|}{\textbf{0.839}} & \multicolumn{1}{c|}{\textbf{0.183}} & \multicolumn{1}{c|}{\textbf{0.915}} & \multicolumn{1}{c|}{\textbf{0.541}}   & \multicolumn{1}{c|}{\textbf{0.570}}     & \multicolumn{1}{c|}{\textbf{0.754}} & \multicolumn{1}{c|}{0.927}                 & 0.621                       \\
\textit{GPT-4o}          & \multicolumn{1}{c|}{0.346}          & \multicolumn{1}{c|}{0.020}          & \multicolumn{1}{c|}{0.783}          & \multicolumn{1}{c|}{0.15}             & \multicolumn{1}{c|}{0.043}              & \multicolumn{1}{c|}{0.363}          & \multicolumn{1}{c|}{\textbf{0.929}}        & \textbf{0.737}              \\
\textit{GPT-4}           & \multicolumn{1}{c|}{0.228}          & \multicolumn{1}{c|}{0.015}          & \multicolumn{1}{c|}{0.791}          & \multicolumn{1}{c|}{0.167}            & \multicolumn{1}{c|}{0.035}              & \multicolumn{1}{c|}{0.320}          & \multicolumn{1}{c|}{0.912}                 & 0.674                       \\
\textit{GPT-3.5}         & \multicolumn{1}{c|}{0.117}          & \multicolumn{1}{c|}{0.017}          & \multicolumn{1}{c|}{0.772}          & \multicolumn{1}{c|}{0.243}            & \multicolumn{1}{c|}{0.028}              & \multicolumn{1}{c|}{0.448}          & \multicolumn{1}{c|}{0.796}                 & 0.568                       \\
\textit{Llama3.1-ins-8b} & \multicolumn{1}{c|}{0.143}          & \multicolumn{1}{c|}{0.015}          & \multicolumn{1}{c|}{0.424}          & \multicolumn{1}{c|}{0.263}            & \multicolumn{1}{c|}{0.009}              & \multicolumn{1}{c|}{0.796}          & \multicolumn{1}{c|}{0.521}                 & 0.593                       \\
\textit{Llama3-ins-8b}   & \multicolumn{1}{c|}{0.189}          & \multicolumn{1}{c|}{0.022}          & \multicolumn{1}{c|}{0.359}          & \multicolumn{1}{c|}{0.285}            & \multicolumn{1}{c|}{0.015}              & \multicolumn{1}{c|}{0.489}          & \multicolumn{1}{c|}{0.459}                 & 0.649                       \\
\textit{Llama2-7b-chat}  & \multicolumn{1}{c|}{0.159}          & \multicolumn{1}{c|}{0.033}          & \multicolumn{1}{c|}{0.207}          & \multicolumn{1}{c|}{0.002}            & \multicolumn{1}{c|}{0.017}              & \multicolumn{1}{c|}{0.398}          & \multicolumn{1}{c|}{0.335}                 & 0.387                       \\
\textit{Chatglm3-6b}     & \multicolumn{1}{c|}{0.011}          & \multicolumn{1}{c|}{0.033}          & \multicolumn{1}{c|}{0.526}          & \multicolumn{1}{c|}{0.000}            & \multicolumn{1}{c|}{0.033}              & \multicolumn{1}{c|}{0.393}          & \multicolumn{1}{c|}{0.403}                 & 0.242                       \\
\textit{Chatglm2-32k-7b} & \multicolumn{1}{c|}{0.039}          & \multicolumn{1}{c|}{0.033}          & \multicolumn{1}{c|}{0.324}          & \multicolumn{1}{c|}{0.000}            & \multicolumn{1}{c|}{0.028}              & \multicolumn{1}{c|}{0.198}          & \multicolumn{1}{c|}{0.212}                 & 0.222                       \\
\textit{Vicuna-v1.5-16k} & \multicolumn{1}{c|}{0.133}          & \multicolumn{1}{c|}{0.013}          & \multicolumn{1}{c|}{0.215}          & \multicolumn{1}{c|}{0.000}            & \multicolumn{1}{c|}{0.015}              & \multicolumn{1}{c|}{0.128}          & \multicolumn{1}{c|}{0.100}                 & 0.377                       \\
\textit{Qwen2-7b-ins}    & \multicolumn{1}{c|}{0.165}          & \multicolumn{1}{c|}{0.026}          & \multicolumn{1}{c|}{0.480}          & \multicolumn{1}{c|}{0.115}            & \multicolumn{1}{c|}{0.048}              & \multicolumn{1}{c|}{0.265}          & \multicolumn{1}{c|}{0.468}                 & 0.617                       \\
\textit{Vicuna-v1.5-7b}  & \multicolumn{1}{c|}{0.180}          & \multicolumn{1}{c|}{0.015}          & \multicolumn{1}{c|}{0.222}          & \multicolumn{1}{c|}{0.002}            & \multicolumn{1}{c|}{0.026}              & \multicolumn{1}{c|}{0.233}          & \multicolumn{1}{c|}{0.383}                 & 0.014                      \\ \hline
	\end{tabular}
      \caption{Performance of graph reasoning.}
    \label{tab:9}
\end{table*}

\begin{table*}[htbp]
\centering
\resizebox{\linewidth}{!}{
\begin{tabular}{p{4cm} p{2.5cm} p{3cm} p{2cm}}
\toprule
\textbf{Level 3} & \textbf{Task} & \textbf{Dataset} & \textbf{Graph \#} \\
\midrule
\multirow{3}{*}{Attribute Understanding} & NN  & BG/TSG/GTG/SPG  & 1840 \\
                                         & AD  & BG/TSG/GTG/SPG  & 1840 \\
                                         & CT  & BG/TSG/GTG/SPG  & 1840 \\
\midrule
Graph Memory                            & MS  & BG/TSG/GTG/SPG  & 1840 \\
\midrule
\multirow{2}{*}{Graph Recognition}       & TR  & Tree Structure Graph  & 460 \\
                                         & BR  & Bipartite Graph  & 460 \\
\midrule
Graph Traversaling                      & BFS  & Graph Traversal Graph  & 460 \\
\midrule
\multirow{3}{*}{Graph QA and Querying}   & NQ  & IMDBText  & 500 \\
                                         & RQ  & IMDBText  & 500 \\
                                         & RN  & IMDBText  & 500 \\
\midrule
Subgraph Extraction                     & SE  & IMDBText  & 500 \\
\midrule
\multirow{3}{*}{\begin{tabular}[c]{@{}l@{}}Simple Graph\\ Theory Problems\end{tabular}} & SP  & Shortest Path Graph  & 460 \\
                                         & MF  & Max Flow Graph  & 460 \\
                                         & EP  & Eulerian Graph  & 460 \\
\midrule
\multirow{3}{*}{\begin{tabular}[c]{@{}l@{}}Complex Graph\\ Theory Problems\end{tabular}} & HC  & Hamiltonian Graph  & 460 \\
                                         & TSP  & TSP Graph  & 460 \\
                                         & GC  & Graph Coloring  & 460 \\
\midrule
Node Entity Reasoning                   & NC  & ACMText  & 500 \\
\midrule
Link Relationship Reasoning             & LP  & ACMText  & 500 \\
\bottomrule
\end{tabular}}
\caption{Overview of Tasks and Datasets for Graph-based Problems.}
\label{tab:10}
\end{table*}

\end{document}